\documentclass{article}
\usepackage[utf8]{inputenc}
\usepackage{authblk}
\usepackage{setspace}
\usepackage[margin=1.25in]{geometry}
\usepackage{graphicx}
\graphicspath{ {./figures/} }
\usepackage{subcaption}
\usepackage{amsmath}
\usepackage{lineno}
\usepackage{color}
\usepackage{CJKutf8}

\usepackage{tabularx} % for 'tabularx' env. and 'X' col. type
\usepackage{ragged2e} % for \RaggedRight macro
\usepackage{booktabs} % for \toprule, \midrule etc macros
\usepackage{longtable}
\usepackage{makecell}
\newcolumntype{L}{>{\RaggedRight\hangafter=1\hangindent=0em}X}

\usepackage{multirow}
\usepackage{supertabular}
\usepackage{booktabs}

\usepackage[style=nejm, 
citestyle=numeric-comp,
sorting=none]{biblatex}
\title{Prospective Role of Foundation Models in Advancing Autonomous Vehicles}

\author[1]{Jianhua Wu}
\author[1,7]{Bingzhao Gao}
\author[1]{Jincheng Gao}
\author[1]{Jianhao Yu}
\author[1*]{Hongqing Chu}

\author[2]{Qiankun Yu}

\author[3]{\\Xun Gong}
\author[3]{Yi Chang}

\author[4]{H. Eric Tseng}
\author[5,6*]{Hong Chen}
\author[6,7]{Jie Chen}

\affil[1]{School of Automotive Studies, Tongji University, Shanghai 201804, China.}
\affil[2]{SAIC Intelligent Technology, Shanghai 201805, China.}
\affil[3]{College of Artificial Intelligence, Jilin University, Changchun 130012, China.}
\affil[4]{Research \& Advanced Engineering, Ford Motor Company, Dearborn, MI 48124 USA (Retired).}
\affil[5]{College of Electronic and Information Engineering, Tongji University, Shanghai 201804, China.}
\affil[6]{National Key Laboratory of Autonomous Intelligent Unmanned Systems, Shanghai 201210, China.}
\affil[7]{Frontiers Science Center for Intelligent Autonomous Systems, Tongji University, Shanghai 201210, China.}

\affil[*]{Address correspondence to: chuhongqing@tongji.edu.cn, chenhong2019@tongji.edu.cn}

\date{}

\begin{document}
\begin{CJK}{UTF8}{gbsn}

\maketitle

%%%%%% Abstract %%%%%%
\begin{abstract}
With the development of artificial intelligence and breakthroughs in deep learning, large-scale Foundation Models (FMs), such as GPT, Sora, etc., have achieved remarkable results in many fields including natural language processing and computer vision. The application of FMs in autonomous driving holds considerable promise. For example, they can contribute to enhancing scene understanding and reasoning. By pre-training on rich linguistic and visual data, FMs can understand and interpret various elements in a driving scene, and provide cognitive reasoning to give linguistic and action instructions for driving decisions and planning. Furthermore, FMs can augment data based on the understanding of driving scenarios to provide feasible scenes of those rare occurrences in the long tail distribution that are unlikely to be encountered during routine driving and data collection. The enhancement can subsequently lead to improvement in the accuracy and reliability of autonomous driving systems. Another testament to the potential of FMs' applications lies in World Models, exemplified by the DREAMER series, which showcases the ability to comprehend physical laws and dynamics. Learning from massive data under the paradigm of self-supervised learning, World Model can generate unseen yet plausible driving environments, facilitating the enhancement in the prediction of road users' behaviors and the off-line training of driving strategies. In this paper, we synthesize the applications and future trends of FMs in autonomous driving. By utilizing the powerful capabilities of FMs, we strive to tackle the potential issues stemming from the long-tail distribution in autonomous driving, consequently advancing overall safety in this domain.
\end{abstract}

%%%%%% Main Text %%%%%%

\section{Introduction}

Autonomous driving, as one of the most challenging tasks in artificial intelligence, has received considerable attention. The conventional autonomous driving system adopts a modular development strategy\cite{yurtsever2020survey,grigorescu2020survey}, whereby perception, prediction, and planning are developed separately and integrated into the vehicle. However, the information transmitted between modules is limited, and there is missing information. Furthermore, there are cumulative errors in the propagation process, and the computational efficiency of the modular transmission is relatively low. These factors collectively result in poor model performance. To further reduce the error and improve the computational efficiency, in recent years, researchers have attempted to train the model in an end-to-end manner\cite{96,chib2023recent}. End-to-end means that the model takes inputs directly from the sensor data and then outputs control decisions for the vehicle directly. While some progress has been made, the models still mainly rely on supervised learning with manually labeled data. Due to the ever-changing driving scenarios in the real world, it is challenging to cover all potential situations with only limited labeled data. This results in a model with poor generalization ability, which makes it difficult to adapt to the complex and changeable real-world driving corner cases.

In recent years, the emergence of Foundation Models (FMs) has provided new ideas to address this gap. A Foundation Model is commonly perceived as a large-scale machine learning model trained on diverse data, capable of being applied to various downstream tasks, which might not necessarily be directly related to its original training objective. The term was coined by Stanford University in August 2021 as ``any model that is trained on broad data (generally using self-supervision at scale) that can be adapted (e.g., fine-tuned) to a wide range of downstream tasks" \cite{1}. Examples of FMs include BERT\cite{2} and the GPT-4\cite{43} in Natural Language Processing (NLP), and Sora\cite{videoworldsimulators2024} in Computer Vision (CV). Most FMs are constructed based on pre-existing architectures. For example, BERT and GPT-4 are based on the Transformer\cite{5}, and Sora is founded on the Diffusion Transformer\cite{peebles2023scalable}.

Different from traditional deep learning, FMs can learn directly from massive unlabeled data (e.g., videos, images, natural language, etc.) through self-supervised pre-training, thereby acquiring stronger generalization ability and emergent abilities (thought to have already appeared in Large Language Models). Based on this, after fine-tuning with a small amount of supervised data, FMs can be rapidly adapted and migrated to downstream tasks such as autonomous driving. With the strong comprehension, inference, and generalization ability imparted by self-supervised pre-training, FMs are expected to break the bottleneck of traditional models, enabling the autonomous driving system to better understand and adapt to complex traffic environments, thus providing a safer and more reliable autonomous driving experience.

\subsection{Emergent Abilities}

Along with FM, \cite{1} talks about the emergence characteristic or emergent ability of FM as ``An ability is emergent if it is not present in smaller models but is present in larger models.” For instance, the adaptability of a Language Model (LM) to diverse downstream tasks, a novel behavior not directly tied to its initial training, seems to emerge abruptly as the model scales beyond an undisclosed threshold, transforming into a Large Language Model (LLM) \cite{9}. 

\begin{figure}[h]
    \centering
    \includegraphics[scale=0.18]{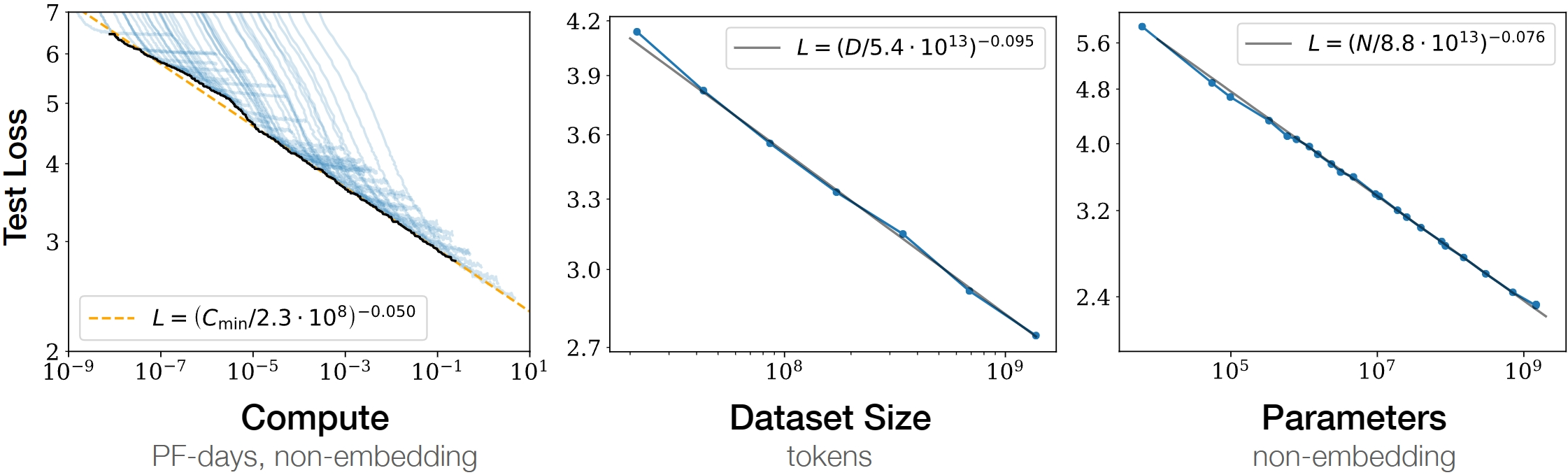}
    \caption{Scaling Laws\cite{10}}
    \label{fig:1}
\end{figure}

\begin{figure}[h]
    \centering
    \includegraphics[scale=0.21]{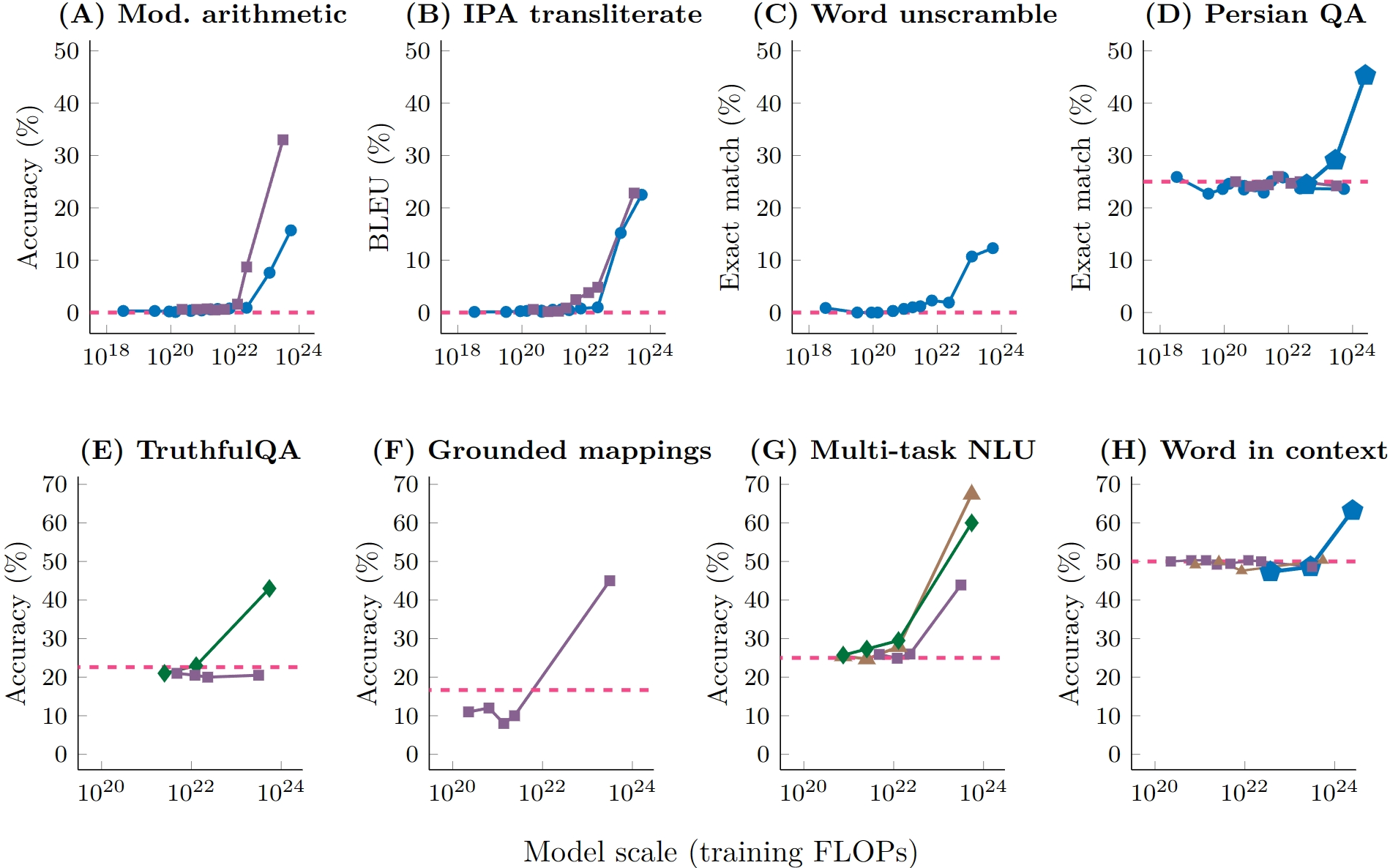}
    \caption{Emergent abilities of LLMs\cite{9}}
    \label{fig:2}
\end{figure}

Currently, the emergent abilities of FMs are mainly reflected in LLMs. In Fig.~\ref{fig:1}\cite{10} it is illustrated that as the model size, dataset size, and the number of computational floats used for training increase, the loss of LLM decreases, providing support for performing large-scale model training. Fig.~\ref{fig:2}\cite{9} shows that when the amount of parameters of the model reaches a certain level, the capabilities of  LLMs will get a qualitative leap, showing emergent abilities in different tasks.

The emergent abilities of LLMs are well represented in In Context Learning (ICL)\cite{9,11}, which, strictly speaking, can be regarded as a subclass of prompt tuning. Context learning ability is the capability of LLMs to learn in a specific contextual environment. The main idea is to learn from analogies\cite{12}. ICL or prompt learning enables LLMs to get excellent performance in a specific context without parameter tuning.

One particular type of ICL is Chain-of-Thought (CoT). Users can break down complex problems into a series of reasoning steps as input to LLM. In this way, LLM can perform complex reasoning tasks\cite{13}. Emergent abilities are commonly found in LLMs, there is currently no compelling explanation for why these abilities would appear the way they do.

Park et al.\cite{14} introduced generative agents that simulated real human behaviors, performed daily activities based on pre-input settings, and stored daily memories in natural language. The authors connected generative agents to LLM to create a small society with 25 intelligent agents, retrieved memories with LLM, and used its emergent abilities to plan the behaviors of intelligent agents. In the experiment, the intelligent agents emerged with more and more social behaviors in addition to their behaviors, fully demonstrating the LLM's intelligent emergence.

\subsection{Pre-training}

The implementation of FMs is based on transfer learning and scaling\cite{1}. The idea of transfer learning\cite{15,16} is to apply the knowledge learned in one mission to another. In deep learning, transfer learning is implemented in two stages, pre-training and fine-tuning. FMs are pre-trained with massive data. After obtaining the pre-trained model a specific dataset is selected for fine-tuning to adapt to different downstream tasks.

Pre-training is the foundation for FMs to obtain emergent abilities. By being pre-trained with massive data, FMs can obtain basic understanding and generative capability. Pre-training tasks include supervised learning (SL), self-supervised learning (SSL), etc\cite{17}. Early pre-training relied on supervised learning, especially in computer vision. To meet the training needs of neural networks, some large-scale supervised datasets, such as ImageNet\cite{18}, were built. However, supervised learning also has some drawbacks, i.e., large-scale data labeling is required. With the gradual increase in the size of the model and the amount of parameters, the drawbacks of supervised learning become more obvious. In NLP, since the degree of difficulty in labeling text is much greater than that of labeling images, SSL is gradually favored by scholars due to its feature of not requiring labeling.

\subsubsection{Self-Supervised Learning}

Self-supervised learning allows learning feature representations in unlabeled data for subsequent tasks. The distinguishing feature of SSL is that they do not require manually labeled labels, but instead generate labels automatically from unlabeled data samples.

SSL usually involves two main processes\cite{19}: Self-supervised training phase: the model is trained to solve a designed pretext task and automatically generates pseudo labels based on data properties in this phase. It is designed to allow the model to learn the generic representation of the data. Downstream tasks application phase: after the self-supervised training, the knowledge learned by the model can be further used for actual downstream tasks. Downstream tasks use supervised learning methods, which include semantic segmentation\cite{20}, target detection\cite{21}, and sentiment analysis\cite{22}, etc. Due to self-supervised training, the generalization ability and convergence speed of the model in downstream tasks will be greatly improved.

SSL methods generally fall into three categories \cite{23}: generative-based, contrastive-based, and adversarial-based. Generative-based method: first encodes the input data using an encoder and then uses a decoder to regain the original form of the data. The model learns by minimizing the error. And generative-based methods include auto-regressive models, auto-encoding models, etc\cite{24}. Contrastive-based method: constructs positive and negative samples by pretext tasks and learns by comparing the similarity with positive and negative samples. Such methods include SimCLR\cite{25} and others. Adversarial-based method: this method consists of a generator and a discriminator. The generator is responsible for generating fake samples, while the discriminator is adapted to distinguishing between these fake samples and real samples\cite{23}, and a typical example is generative adversarial network\cite{26}.

\subsubsection{Pretext Tasks of Self-Supervised Learning}

The pretext tasks can also be referred to as self-supervised tasks as they rely on the data itself to generate labels. These tasks are designed to make the model learn representations that are relevant to a specific task, thereby better handling downstream tasks.

In computer vision, the method of designing pretext tasks according to data attributes includes four main categories\cite{19}: generation-based, context-based, free semantic label-based, and cross-modal-based. Among them, generation-based approaches mainly involve image or video generation tasks\cite{27,28}; context-based pretext tasks are mainly designed leveraging contextual features of images or videos, such as contextual similarity, spatial structure, temporal structure, etc.\cite{29,30,31}; in the free semantic label-based pretext tasks, the network is trained leveraging automatically generated semantic labels\cite{32}; and cross modal-based pretext tasks need to consider multiple modalities such as vision and voice\cite{33}.

In natural language processing, the most common pretext tasks include\cite{34}: center and neighbor word prediction, next and neighbor sentence prediction, autoregressive language modeling, sentence permutation, masked language modeling, etc. Among them, the Word2Vec\cite{35} model uses center word prediction as a pretext task, while the BERT model uses next sentence prediction and masked language modeling as pretext tasks. These models are trained to learn the expressions of the corpus and applied to downstream tasks.

\subsection{Fine-tuning}

Fine-tuning is the process of further training on a specific task based on an already trained model, to adapt it to the specific data and requirements of the task. Typically, a model that has been pre-trained on large-scale data is used as a foundational model, and then it is fine-tuned on a specific task to improve performance. Currently, in the field of LLMs, fine-tuning methods include two main approaches: instruction tuning and alignment tuning\cite{36}.

Instruction fine-tuning aims at fine-tuning pre-trained models on a collection of datasets described by instructions\cite{37}. Instruction fine-tuning generally includes two phases. First, instances of instruction formatting need to be collected or created. And then these instances are used to fine-tune the model. Instruction fine-tuning allows LLMs to exhibit strong generalization ability on previously unseen tasks. The models obtained after pre-training and fine-tuning can work well in most cases, however, some special cases may occur. In the case of LLM, for example, the trained model may appear to fabricate false information or retain biased information from the corpus. To avoid such problems, the concept of human-aligned fine-tuning was proposed. The goal is to make the model's behaviors conform to human expectations\cite{38}. In contrast to instruction fine-tuning, this kind of alignment requires the consideration of completely different standards.

The GPT family is a typical FM, and its training process also includes pre-training and fine-tuning. Taking ChatGPT as an example, the pre-training process of ChatGPT uses self-supervised pre-training \cite{3}. Given an unsupervised corpus, a standard language modeling approach is used to optimize its Maximum Likelihood Estimate (MLE). GPT uses a multi-layer transformer decoder architecture\cite{39},  resulting in a pre-trained model.

The fine-tuning phase of ChatGPT consists of the following three steps\cite{38}. Firstly, supervised fine-tuning (SFT) is performed on the obtained pre-trained model; secondly, comparison data are collected to train the reward model; and thirdly, the SFT model is fine-tuned to maximize the reward leveraging the PPO algorithm\cite{40}. The last two steps together are Reinforcement Learning with Human Feedback (RLHF)\cite{41}.

\subsection{Abilities of Foundation Models in Autonomous Driving}

The ultimate goal of autonomous driving is to realize a driving system that can completely replace human driving, and the basic criterion for evaluation is to drive like a human driver, which puts forward very high requirements on the reasoning ability of autonomous driving models. We can see that FMs based on large-scale data learning have powerful reasoning and generalization ability, which have great potential in autonomous driving. FMs can be used to enhance scenarios understanding, give language-guided commands, and generate driving actions in empowering autonomous driving. In addition, FMs can be augmented with powerful generative capability for data augmentation, including extending existing autonomous driving datasets and directly generating driving scenarios. In particular, World Models (a type of FM) can learn the inner workings of the physical world and predict future driving scenarios, which is of significant importance for autonomous driving.

Consequently, it was deemed appropriate to conduct a comprehensive review of the applications of FMs in autonomous driving. This paper provides that review.
\begin{itemize}
	\item In Section 2, a brief overview of the latest supervised end-to-end autonomous driving is provided, to offer the reader a better background understanding. 
	\item Section 3 reviews the applications of language and vision FMs in enhancing autonomous driving.
	\item Section 4 reviews the applications of World Models in the exploration of the field of autonomous driving.
        \item Section 5 reviews the applications of FMs in data augmentation.
\end{itemize}

Building on the preceding overview, Section 6 presents the challenges and future directions for enhancing autonomous driving with FMs.

\section{Supervised End-to-end Autonomous Driving}

\begin{figure}[h]
    \centering
    \includegraphics[scale=0.52]{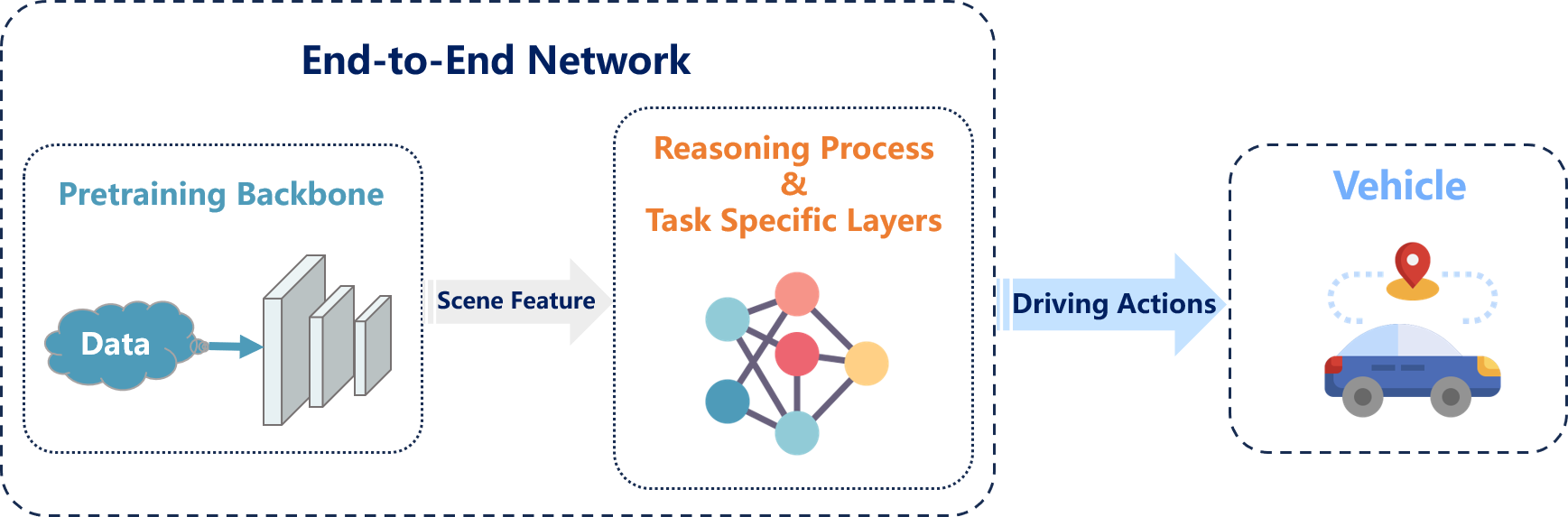}
    \caption{The pipeline diagram for the supervised end-to-end autonomous driving system with a Pretraining Backbone. Multi-modal sensing information is input to the Pretraining Backbone to extract features, after which it enters into the framework of autonomous driving algorithms built by various methods to realize tasks, such as planning/control, to accomplish end-to-end autonomous driving tasks.}
    \label{fig:4}
\end{figure}

The research idea of ``pre-training + fine-tuning'' in autonomous driving research has not only appeared after the introduction of large models but has been researched for a long time. To use a more familiar term, it is end-to-end autonomous driving. In the past few years, some scholars have already optimized Pretraining Backbone in various ways, including the transformer architecture and self-supervised learning methods——Note that we define Pretraining Backbone here refers to a model that transforms each modal input into a usable feature representation for downstream tasks (such as target detection, trajectory prediction, decision planning, etc.). Many research attempts have also been made to develop end-to-end frameworks based on the transformer architecture, with excellent results. Hence, to summarize the application of the underlying models in autonomous driving more comprehensively, we believe that it is necessary to introduce the research related to end-to-end autonomous driving based on Pretraining Backbone. In this section, we summarize the latest research on Pretraining Backbone with end-to-end autonomous driving solutions. The pipeline for such methods is briefly illustrated in Fig.~\ref{fig:4}.

\subsection{Pretraining Backbone}

In end-to-end modeling, feature extraction of low-level information from raw data determines the potential of subsequent model performance to a certain extent, and an excellent Pretraining Backbone can endow the model with more powerful feature learning capability.

Pretraining convolutional networks such as ResNet\cite{79}, and VGG\cite{80} are the most widely used backbone for visual feature extraction in end-to-end models. These pretraining networks are often trained to leverage target detection or segmentation as the task to extract generalized feature information, and their competitive performances have been verified in many works. ViT\cite{6} first applied the transformer architecture to image processing and achieved excellent classification results. Transformer has the advantage of optimized algorithms for handling large-scale data with its simpler architecture and faster inference speed. The self-attention mechanism is very suitable for processing time series data, there is the possibility of modeling and predicting the temporal motion trajectories of objects in the environment, and is conducive to the fusion of heterogeneous data from multiple sources, such as LiDAR point clouds, images, maps, etc.

Another class of Pretraining Backbone, represented by LSS\cite{81}, BEVDet\cite{82}, BEVformer\cite{83}, BEVerse\cite{84}, etc., expanded the usability by extracting the images captured by the surround-view camera and converting them to Bird's-Eye View (BEV) features through model learning, indexing the local image features from the 2D-viewpoints to the 3D-space. In recent years, BEV has attracted extensive interest due to its ability to describe the driving scene more accurately, and the research of leveraging BEV features such as Pretraining Backbone output is not limited to camera, and the extraction and fusion of multi-modal sensing BEV features represented by BEVfusion\cite{85} has further provided a wider vision for autonomous driving systems. However, it should be pointed out that although the transformer architecture brings great performance enhancement, these Backbone still constructs pre-trained models with supervised learning methods, which rely on massive labeled data, and the data quality also greatly affects the final result of the model.

In both camera and point cloud processing domains, some works implement the Pretraining Backbone by unsupervised or self-supervised learning methods. Wu et al.\cite{86} proposed the PPGeo model, which uses a large number of unlabeled driving videos to accomplish the pre-training of the visual coder in two stages, and can be adapted to different downstream end-to-end autonomous driving tasks. Sautier et al.\cite{87} proposed BEVContrast for self-supervision of 3D Backbone on automotive LiDAR point clouds, which defines contrasts at the level of 2D Cells in the BEV plane, retaining the simplicity as in PointContrast\cite{88} while maintaining good performance in downstream driving tasks. Especially, while the self-supervised learning approach of "masking + reduction" is also considered to be an effective way of modeling the world, Yang et al.\cite{yang2023unipad} proposed UniPAD, which is implemented based on self-supervised learning methods for MAE and 3D rendering. A portion of these multi-modal data is randomly keyed out to be masked and transformed into voxel space, where RGB or depth prediction results are generated by rendering techniques in such a 3D space, and the rest of the original images are used as the generated data for supervised learning. The flexibility of the approach enables good integration into both 2D and 3D frameworks, and downstream tasks such as depth estimation, target detection, segmentation, and many others fine-tuned and trained on the model perform superiorly.

\subsection{Supervised End-to-end Autonomous Driving Models}

Early work on end-to-end autonomous driving modeling was mainly based on various types of deep neural networks, which were constructed through imitation learning\cite{89,90,91,92,hu2022st} or reinforcement learning\cite{93,94,95} methods. The work of Chen et al.\cite{96} analyzed the key challenges facing end-to-end autonomous driving from a methodological perspective, pointing out the future trend of empowering end-to-end autonomous driving with fundamental models such as transformer. Some scholars have tried to build an end-to-end autonomous driving system with transformer and got competitive results. For instance, there have been Transfuser\cite{97,98}, NEAT\cite{99}, Scene Transformer\cite{100}, PlanT\cite{101}, Gatformer\cite{102}, FusionAD\cite{103}, UniAD\cite{104}, VAD\cite{jiang2023vad}, GenAD\cite{zheng2024genad} and a host of end-to-end frameworks developed based on transformer architecture.

Chitta et al.\cite{97,98} proposed Transfuser, which takes RGB images and BEV views from LiDAR as inputs, uses multiple transformers to fuse the feature maps, and predicts the trajectory points for the next four steps through a single-layer GRU network, followed by longitudinal and transverse PIDs to control the vehicle operation. NEAT\cite{99} further mapped the BEV scene to trajectory points and semantics information, then used an intermediate attention map to compress high-dimensional image features, which allows the model to focus on driving-relevant regions and ignore driving task-irrelevant information. PlanT proposed by Renz et al.\cite{101} used simple object-level representations (vehicles, roads) as inputs to the transformer encoder, and used speed prediction of surrounding vehicles as a secondary task to predict future waypoint trajectories. UniAD proposed by Hu et al.\cite{104} enhanced the design of the decoder and achieved the integration of the full stack of autonomous driving tasks into a unified framework to improve autonomous driving performance, although still relying on different sub-networks for each task. This work also won the CVPR 2023 Best Paper Award, which shows the academic recognition of the end-to-end autonomous driving paradigm. However, these models often require intensive computation. For this reason, Jiang et al.\cite{jiang2023vad} proposed a method to fully vectorize the driving scenarios and learn the instance-level structural information to improve computing efficiency. In contrast to the previous modular end-to-end planning, Zheng et al.\cite{zheng2024genad} propose a generative end-to-end, modeling autonomous driving as a trajectory generation.

Moreover, Drive Anywhere proposed by Wang et al.\cite{105} not only realizes end-to-end multi-modal autonomous driving but also combines with LLM to be able to provide driving decisions based on representations that can be queried through images and texts. Dong et al.\cite{106} generated image-based action commands and explanations by building a feature extraction model based on transformer. Jin et al.\cite{107} proposed the ADAPT model to directly output vehicle control signals with inference language descriptions through an end-to-end model. This is the first driving action captioning architecture based on an action-aware transformer. It accomplished the driving control task while adding natural language narratives to guide the decision-making and action process of the autonomous driving control module. It also helped the user to get the vehicle's state and the surrounding environment at all times and to better understand the basis of the actions taken by the autonomous driving system, which improved the interpretability of the decision-making. This provides a glimpse of the potential of the transformer architecture to improve the interpretability of end-to-end driving decisions.

\section{Human-like Driving Based on Language and Vision Models}

With significant research progress in LLMs BERT, GPT-4, Llama\cite{44}, Vision Language Models (VLMs) CLIP\cite{4}, ALIGN\cite{45}, BLIP-2\cite{47} and Multi-modal Large Language Models(M-LLMs) GPT-4V\cite{49}, LLaVA\cite{48} and Gemini\cite{team2023gemini}, as well as other FMs, their powerful reasoning capabilities are considered to have ushered in a new dawn for the realization of Artificial General Intelligence (AGI)\cite{50}, which has had a significant and far-reaching impact on all aspects of society. In autonomous driving, FMs such as language and vision also show great potential, which is expected to improve the understanding and reasoning ability of autonomous driving models on driving scenarios and realize human-like driving for autonomous driving. 

\begin{figure}[h]
    \centering
    \includegraphics[scale=0.58]{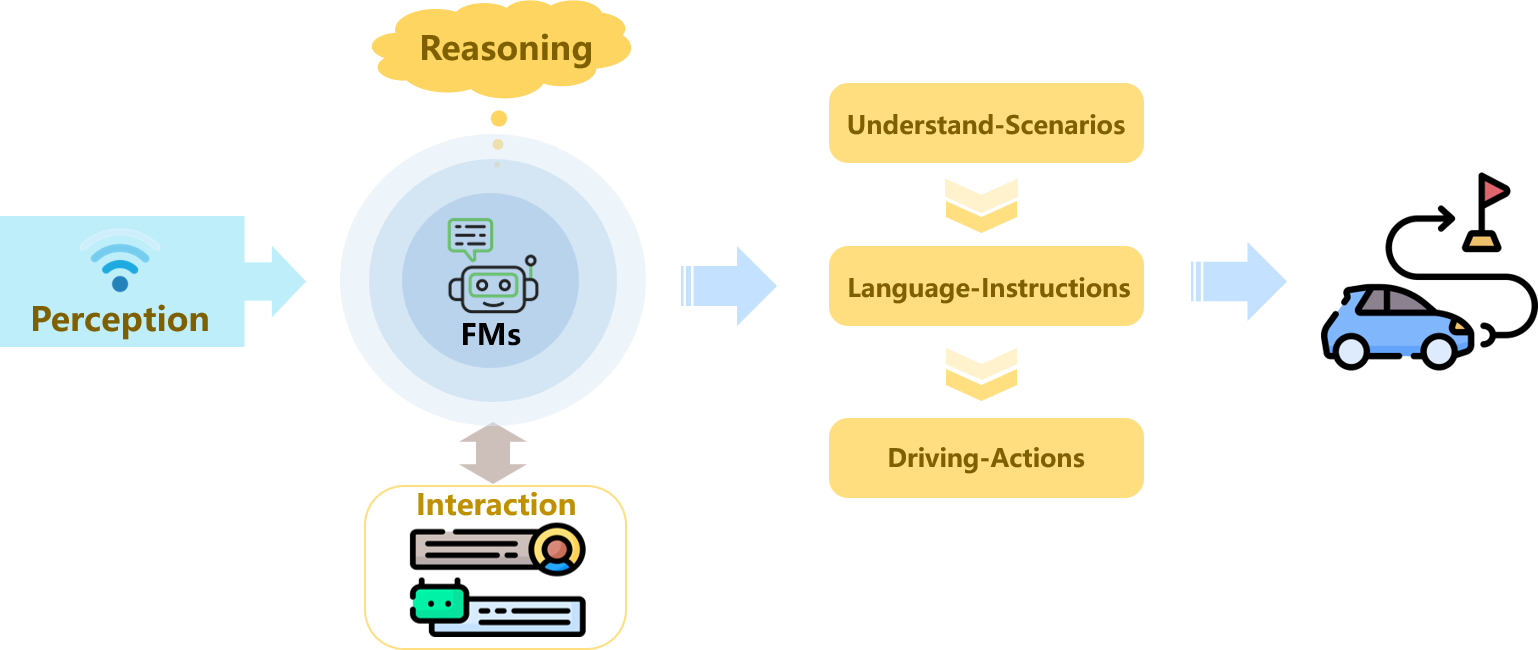}
    \caption{The pipeline diagram for enhancing autonomous driving leveraging FMs, where FMs refer to language models and vision models. FMs can learn perceptual information and utilize their powerful ability to understand the driving scenarios and reason to give language-guided instructions and driving actions to enhance autonomous driving.}
    \label{fig:3}
\end{figure}

We provide an introduction of the research related to the enhancement of the understanding of the driving scenarios by the autonomous driving system based on FMs of language and vision, as well as the reasoning to give language-guided instructions and driving actions, as illustrated in Fig.~\ref{fig:3}. Related work on enhancing the understanding of driving scenarios is presented in Section 3.1, on reasoning to give language-guided instructions in Section 3.2, and on reasoning to generate driving actions in Section 3.3.

\subsection{Understanding of Driving Scenarios}

The study by Vasudeva et al.\cite{51} found that the ability of the model to comprehend the scene and localize objects can be effectively enhanced by acquiring verbal descriptions and gaze estimation, etc. Li et al.\cite{52} proposed an image captioning model that generates high-level semantic information to improve its comprehension of the traffic scene. Their work verified that linguistic and visual features can effectively enhance the comprehension of driving scenarios.

Sriram et al.\cite{54} have proposed an autonomous navigation framework that combines semantic segmentation results with natural language commands. This framework has been verified to be effective as a vehicle driver in the CARLA simulator and the KITTI dataset\cite{126}. Elhafsi et al.\cite{57} identified semantic anomalies by converting observed visual information into natural language descriptions and passing them to the LLM to exploit its powerful reasoning capabilities. In the context of VLMs applications, Chen et al.\cite{53} transferred image and text features to a 3D point cloud network based on CLIP to enhance the model's understanding of the 3D scene. Romero et al.\cite{56} constructed a video analytics system based on VIVA\cite{romero2022optimizing}, an extended model of CLIP, intending to improve query accuracy through the utilization of the powerful comprehension of VLM. Tian et al.\cite{tian2024drivevlm} employed VLM to describe and analyze the driving scenarios, thereby enhancing the understanding of the driving scenarios. In addition to direct enhancement of scene data, perceptual features have also been explored for enhancement. Pan et al.\cite{pan2024vlp} designed the Ego-car prompt to enhance the obtained BEV features using the language model in CLIP. Dewangan et al.\cite{58} proposed an approach to enhance BEV maps by detecting the features of each object in the BEV through VLMs (Blip-2\cite{47}, Minigpt-4\cite{59}, Instructblip\cite{60}) and through linguistic characterizations to obtain a language-enhanced BEV map. However, existing VLMs are constrained to the 2D domain, lacking the capacity for spatial awareness and long-horizon extrapolation. To address this issue, Zhou et al.\cite{zhou2024embodied} proposed a model——Embodied Language Model (ELM), which enhances the understanding of driving scenarios over long-time domains and across space. This is achieved by using diverse pre-training data and selecting adaptive tokens.

\subsection{Language-Guided Instructions}

Here we present a review of studies that give linguistic instructions through FMs, mainly descriptive instructions, such as ``Red light ahead, you should slow down", ``Intersection ahead, please pay attention to pedestrians", etc. Ding et al.\cite{61} used a visual coder to encode video data, which was then fed into a large language model to generate corresponding driving scenario descriptions and suggestions. In particular, this work also proposed a method that enables high-resolution feature maps and the obtained high-resolution information to be fused into M-LLMs to further enhance the model's recognition, interpretation, and localization capabilities. Fu et al.\cite{62} explored the potential of leveraging LLMs to comprehend driving environments like a human being, utilizing the LLaMA-Adapter\cite{63} to describe the scene data, and then giving linguistic commands via GPT-3.5. Wen et al.\cite{64} proposed DiLu, a knowledge-driven paradigm based on previous work that can make decisions based on common-sense knowledge and accumulate experience. In particular, the article pointed out that DiLu possesses the ability to direct experience acquisition of real-world data, which has the potential for practical deployment of autonomous driving systems. To further improve the safety of LLM-based autonomous driving, Wang et al.\cite{wang2023empowering} used an MPC-based verifier to evaluate and provide feedback on trajectory planning, and then fused prompt learning to enable LLM to perform in-context safety learning, which overall improved the safety and reliability of autonomous driving. In order to enrich the data input to obtain more accurate scene information, Wang et al.\cite{wang2023drivemlm} utilized multi-model LLM to enable the autonomous driving system to obtain linguistic commands. Meanwhile, for the gap between linguistic commands and vehicle control commands, this work performed an alignment operation on decision states. 

The aforementioned works are more in the context of data sets and simulation environments, and there has been some exploratory work in terms of real vehicle testing. Wayve proposed LINGO-1\cite{65}, a grand model of self-driving interaction based on a visual-verbal-action grand model, where the model can interpret itself and answer visually while driving. It introduced human driving experiences, which can explain various causal elements in the driving scenarios through natural language descriptions, acquire feature information in the driving scenario in a human-like understanding way, and learn and give interactive language commands. Cui et al.\cite{cui2023large} innovatively placed the LLM in the cloud, entered human commands, and leveraged the reasoning ability of the LLM to generate executable code instructions. However, the work suffers from latency issues and has room for improvement in terms of real-time performance requirements for autonomous driving.

\begin{figure}[h]
    \centering
    \includegraphics[scale=0.45]{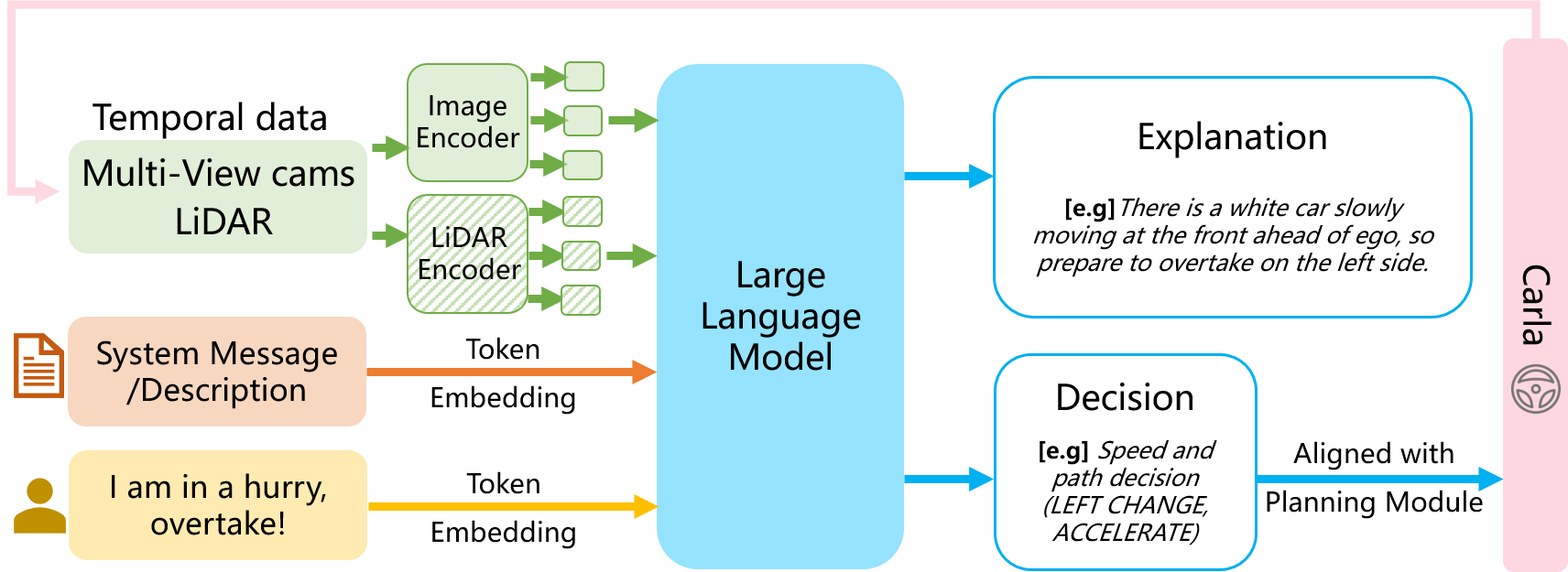}
    \caption{For the application of LLMs to autonomous driving system decision-making, a typical pipeline is shown in this figure, referenced from DriveMLM\cite{wang2023drivemlm}.}
    \label{fig_eg1}
\end{figure}

The pipeline for incorporating LLMs into autonomous driving systems in more current research is summarized in general terms in Fig.\ref{fig:3}, which is mainly implemented through scene understanding, high-level semantic decision-making, and trajectory planning. In this subsection, we summarize the advanced decision-making applications and argue that the research processes have some similarities. To more clearly illustrate how they work, we use DriveMLM\cite{wang2023drivemlm}, a typical recent research work, as an example for further illustration in Fig.\ref{fig_eg1}.

DriveMLM simulates the behavioral planning module of a modular autopilot system by using a Multi-modal LLM (M-LLM), which performs closed-loop autonomous driving in a realistic simulator based on processed perceptual information and command requirements. DriveMLM also generates natural language explanations of its driving decisions, thereby increasing the transparency and trustworthiness of the system.

\subsection{Generation of Actions}

As described in the previous Section 3.2, academia and industry have attempted to embed GPT linguistic knowledge into autonomous driving decisions to enhance the performance of autonomous driving in the form of linguistic instructions to promote the application of FMs to autonomous driving. Long before FMs made breakthroughs in the LLMs field, some works attempted to improve the performance of autonomous driving through similar research ideas. For example, the MP3 framework proposed by Casas et al.\cite{66} used high-level semantic information as a decision training guide, which together with sensory data constitutes the input to build algorithms to realize motion prediction.

The research on the application of LLMs in autonomous driving is on the ascendant, and the GPT series, as the most successful variant of the transformer architecture, may be able to bring breakthroughs to improve the comprehensive performance at multiple levels. LLM is a representative of FMs from the level of linguistic knowledge that empowers the development of autonomous driving, however, linguistic descriptions and reasoning are not directly applied by the autonomous driving system. Considering that the large model is expected to be truly deployed at the vehicle end, it needs to eventually fall on the planning or control instructions, i.e., FMs should eventually empower autonomous driving from the action state level. Nevertheless, how to quantize linguistic decisions into action commands, such as planning and control, available to the autonomous driving system still faces great challenges. Some scholars have already made preliminary explorations, but there is still much room for development. What's more, some scholars have explored the construction of autonomous driving models through a GPT-like approach, which directly outputs trajectories and even control commands based on LLM. In Table \ref{tab:1} we provide a brief overview of some representative works.

Sha et al.\cite{67} proposed LanguageMPC, which employs GPT-3.5 as a decision-making module for complex autonomous driving scenarios that require human common sense comprehension. By designing cognitive pathways for integrated reasoning in LLM, sha et al. proposed algorithms to transform LLM decisions into actionable driving control commands, which improved the vehicle's ability to handle complex driving behaviors. Jain et al.\cite{68} achieved navigation localization and further planning of trajectories with the help of visual perception for explicit verbal commands. Omama et al.\cite{69} constructed a multi-modal map-based navigation and localization method called ALT-Pilot, which can be used to navigate to arbitrary destinations without the need for high-definition LiDAR maps, demonstrating that off-the-shelf visual language models can be used to construct linguistically enhanced terrain maps. Pan et al.\cite{pan2024vlp} proposed the VLP method to improve the contextual reasoning for visual perception and motion planning of an autonomous driving system with the powerful reasoning capability of LLM in the training phase, and achieved excellent performance in the open-loop end-to-end motion planning task.

Some scholars have also attempted to construct autonomous driving models directly through a GPT-like approach, i.e., leveraging LLMs to construct an end-to-end autonomous driving planner, which directly outputs predicted trajectories, path planning, and even control commands, intending to effectively improve the ability of autonomous driving models to generalize to unknown driving scenarios. 

Pallagani et al.\cite{73} constructed Plansformer, which is both a large language model and a planner, showing the great potential of a large language model fine-tuned as a planner from a variety of planning tasks. Wang et al.\cite{74} constructed the BEVGPT model, which takes as input information about the current surroundings on the road and then outputs a sequence that includes future vehicle decision instructions and spatial paths that can be followed by self-driving vehicles. 

Some works\cite{70, drivelm, 75, 76, 77, lmdrive} took both text prompts and information about the current surroundings on the road as inputs and then output textual responses or interpretations and a sequence that includes future vehicle decision instructions and spatial paths that can be followed by the self-driving vehicle. Among them, Cui et al.\cite{76} utilized GPT-4 with inputs of natural language descriptions and environment perception data to make LLM directly output driving decisions and operation commands, and further experimented with highway overtaking and lane changing scenarios in Ref.\cite{77} to compare driving decisions provided by LLM with different cues, and the study showed that chained-thinking cueing helps LLM to make better driving decisions. 

Some scholars have also tried different ideas. Seff et al.\cite{71} proposed MotionLM, which uses motion prediction as a language modeling task to learn multi-modal distributions by representing continuous trajectories as discrete sequences of motion tokens leveraging a single standard language modeling objective to predict the future behaviors of road network participants. Mao et al.\cite{72} proposed the GPT-Driver model to reformulate the motion-planning task as a language modeling problem by representing the inputs and outputs of the planner as linguistic tokens and leveraging the LLM to generate driving trajectories through linguistic descriptions of coordinate positions. Furthermore, they \cite{mao2023language} proposed Agent Driver, which utilized LLM to introduce a general-purpose library of tools accessible via function calls, cognitive memory for common sense and empirical knowledge for decision-making, and a reasoning machine capable of Chain-of-Thought (CoT) reasoning, task planning, motion planning, and self-reflection to achieve a more nuanced, human-like approach to autopilot. Ma et al.\cite{ma2023dolphins} proposed Dolphins. It is capable of performing tasks such as scene understanding, behavior prediction, and trajectory planning. This work demonstrates the ability of a visual language model to provide a comprehensive understanding of complex and open-world long-tail driving scenarios and solve a range of AV tasks, as well as emergent human-like capabilities including context-learning gradient-free immediate adaptation and reflective error recovery.

Considering the scale challenges of the Visual Language Model (VLM), Chen et al.\cite{132}, based on the idea that digital vector modalities are more compact than image data, fused vectorized 2D scene representations with pre-trained LLMs to improve the LLM's ability to interpret and reason about the integrated driving situation, giving scene interpretation and vehicle control commands. Tian et al. \cite{tian2024drivevlm} propose DriveVLM, which, through the CoT mechanism, is not only able to generate descriptions and analyses of the scenes presented in image sequences to make driving decision guidance but also further enables trajectory planning in conjunction with the traditional automated driving pipeline. The proposed work also provides possible solutions to the challenges inherent in VLM in terms of spatial reasoning and computation, realizing an effective transition between existing autopilot approaches and large model-based approaches.

\begin{figure}[h]
    \centering
    \includegraphics[scale=0.4]{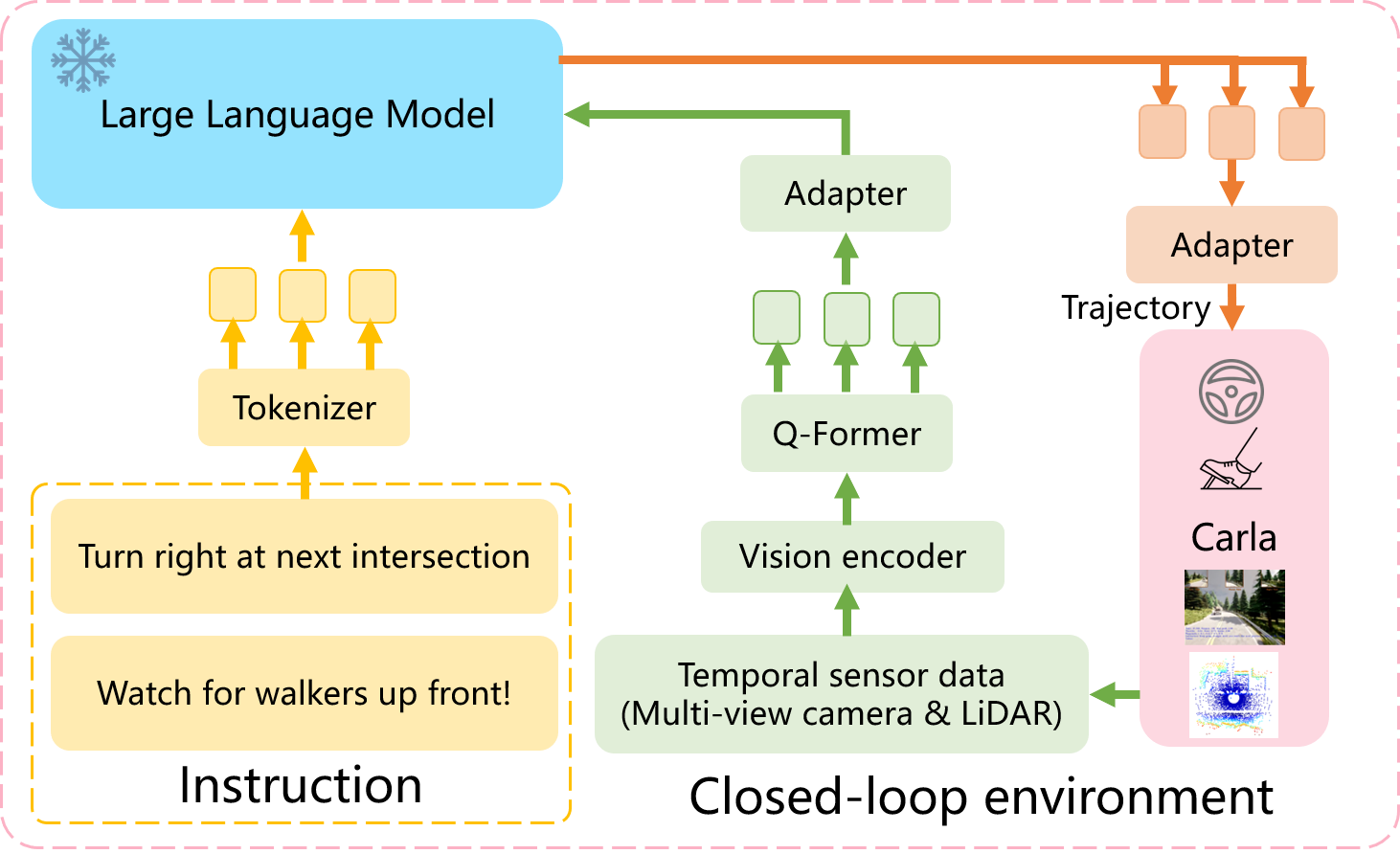}
    \caption{For the application of LLMs to autonomous driving system planning, a typical pipeline is shown in this figure, referenced from LMDrive\cite{lmdrive}.}
    \label{fig_eg2}
\end{figure}

As in the previous subsection, for the research work on the application of LLMs to the direct generation of trajectory planning for autonomous driving systems, we take the example of a typical recent research work in Fig.\ref{fig_eg2}, LMDrive\cite{lmdrive}, to hopefully illustrate more clearly how it works. LMDrive is based on the Carla simulator, and the model training consists of two phases: pre-training and command fine-tuning. In the pre-training phase, prediction headers are added to the vision encoder to perform pre-training tasks. After the pre-training is completed, the prediction headers are discarded and the vision encoder is frozen. In the instruction fine-tuning stage, the navigation instruction and the notice instruction are configured for each driving segment, and the visual tokens are processed through the time series of instruction encoding by LLaMA, and together with the textual tokens, they are inputted into LLM to obtain the prediction tokens. After the 2-MLP Adapter, the output is the planning of the future trajectory of the auto-vehicle and the flag of whether the instruction is completed or not, and the planned trajectory completes the closed-loop simulation through the transverse and longitudinal PID controllers.

This type of research idea is much closer to human driving than pure knowledge embedding to make an autonomous driving model. With the development of large models, it perhaps has the potential to become one of the main development directions in the future. Motion planning, as one of the fundamental topics in the field of intelligent robotics\cite{wulker2019quantizing}, is significant to quantifying linguistic decisions into action commands such as planning and even control available for autonomous driving systems through LLM. However, it should be noted that these new frameworks are also questionable in terms of reliability due to the unresolved pitfalls of large models themselves, such as ``illusions" (LLMs may generate content that conflicts with source or factual information). Specific details about the problems of the large models themselves and the challenges inherited in autonomous driving will be discussed in detail in Section 6.

\begin{center}
\begin{longtable}[h]{
         >{\raggedright\arraybackslash}m{1.3cm}
         >{\raggedright\arraybackslash}m{1.8cm}
         >{\raggedright\arraybackslash}m{1.7cm}
         >{\raggedright\arraybackslash}m{1.2cm}
         >{\raggedright\arraybackslash}m{7.0cm}
        }
    \caption{Works on the use of LLMs for generating autonomous driving planning and control}
    \label{tab:1} \\

            \hline
            Authors & Input & Output & Learning & Description \\  
             \hline
            \makecell[l]{Sha\\et al.\cite{67}\\2023} & Prompt of scenario & Control Action Sequences & \makecell[c]{SL} & This work enables navigation and localization as well as further trajectory planning through visual perception and verbal commands.\\
             \hline
            \makecell[l]{Omama\\et al.\cite{69}\\2023} & OSM Maps with Descriptions, LiDAR, Camera & Location, Trajectory  & \makecell[c]{SL} & This work utilizes a Bayesian state estimation model leveraging visual-linguistic features to generate global paths, plan trajectories, and control vehicles to complete navigation.\\
             \hline
            \makecell[l]{Keysan\\et al.\cite{70}\\2023} & Scene Raster, Text Prompt & Trajectory & \makecell[c]{SL} & This work encodes the driving scene and text prompt with pre-trained models dedicated for each modality, finally sifts through the set of trajectories to find the target trajectory.\\
             \hline
            \makecell[l]{Seff\\et al.\cite{71}\\2023} & Multi-modal Scene & Motion Prediction & \makecell[c]{SL} & This work uses a single standard linguistic modeling objective to learn multi-modal distributions for predicting the future behaviors of traffic participants.\\
             \hline
            \makecell[l]{Mao\\et al.\cite{72}\\2023} & Perception, Ego-States, Trajectory, Goal & Thoughts, Driving Decisions, Trajectory & \makecell[c]{SL} & This work represents the inputs and outputs as linguistic tokens and utilizes the LLM to generate driving trajectories and provide explanations for decision making.\\
             \hline
            \makecell[l]{Wang\\et al.\cite{74}\\2023} & Scene Information & Driving Decisions, Trajectory & \makecell[c]{SL} & In the pre-training stage, this work trains a causal transformer for driving scenario prediction and decision-making. In the fine-tuning stage, it adapts to motion planning and accurate BEV generation.\\
             \hline
            \makecell[l]{Xu\\et al.\cite{75}\\2023} & Video, Text Prompt, Control Signal& Decision, Control Signal & \makecell[c]{SL} & This work tokenizes video sequences, text, and control signals to build the model, which can generate responses to human inquiries and predict control signals.\\
            \hline
            \makecell[l]{Sima\\et al.\cite{drivelm}\\2023} & Video, Text Question& Scene Description, Decision, Trajectory & \makecell[c]{SL} & Based on Graph Visual Question Answering (GVQA), this work realizes structured reasoning for perception, prediction, and planning through suitable quizzes for human-like autonomous driving.\\
            \hline
            \makecell[l]{Shao\\et al.\cite{lmdrive}\\2023} & Camera, LiDAR, Text Prompt& Trajectory & \makecell[c]{SL} & This work accomplishes e2e autonomous driving by interacting with dynamic environments through multi-modal multi-view sensor data and language commands.\\
            \hline
            \makecell[l]{Ma\\et al.\cite{ma2023dolphins}\\2023} & Video, Text Prompt, Control Signal& Scene Description, Prediction, Trajectory & \makecell[c]{SL} & This work employs a Grounde-CoT process to enhance the model's reasoning capabilities. This work also integrates four different tasks to facilitate the model's comprehensive understanding of complex driving scenarios.\\
             \hline
\end{longtable}
\end{center}

\section{Prediction of Autonomous Driving Based on World Models}

World Models refer to the mental models of the world. It can be interpreted as a type of artificial intelligence model that encompasses a holistic understanding or representation of the environment in which it operates. This model is capable of simulating the environment to make predictions or decisions. The term ``World Models" has been mentioned in connection with reinforcement learning in recent literature\cite{137,levine2021understanding}. This concept has also gained attention in autonomous driving because of its capacity to comprehend and articulate the dynamics of the driving environment, as will be detailed below. In his position paper, LeCun\cite{135} pointed out that the learning capability of humans and animals may be rooted in their capacity to learn World Models, allowing them to internalize and understand how the world works. He pointed out that humans and animals have exhibited the ability to acquire a vast amount of background knowledge about the functioning of the world observing a small number of events, whether related or unrelated to the task at hand. The idea of World Model can be traced back to Dyna, proposed by Sutton et al.\cite{136} in 1991, to observe the state of the world and take appropriate actions accordingly to learn interactively with the world\cite{fan2023advanced}. Dyna is essentially a form of reinforcement learning under supervised conditions. After that, researchers have also made many attempts. Ha et al.\cite{137} attempted to learn by leveraging an unsupervised approach, VAE to encode input features, and RNN to learn the evolution of the state. Hafner et al.\cite{138} proposed the Recurrent State Space Model (RSSM), which combined reinforcement learning to realize multi-step prediction that integrates stochasticity and determinism. Based on the RSSM architecture, Hafner et al. successively proposed DreamerV1\cite{139}, DreamerV2\cite{140}, and DreamerV3\cite{141}, which learned in implicit variables to realize image prediction generation. Gao et al.\cite{142} considered that there was redundant information implicit, and extended the framework of the Dreamer series by proposing the Semantic Masked recurrent World Model (SEM2) to learn relevant driving states. Hu et al.\cite{143} removed prediction rewards and proposed an imitation learning-based method MILE to predict future states.

\begin{figure}[h]
    \centering
    \includegraphics[scale=0.51]{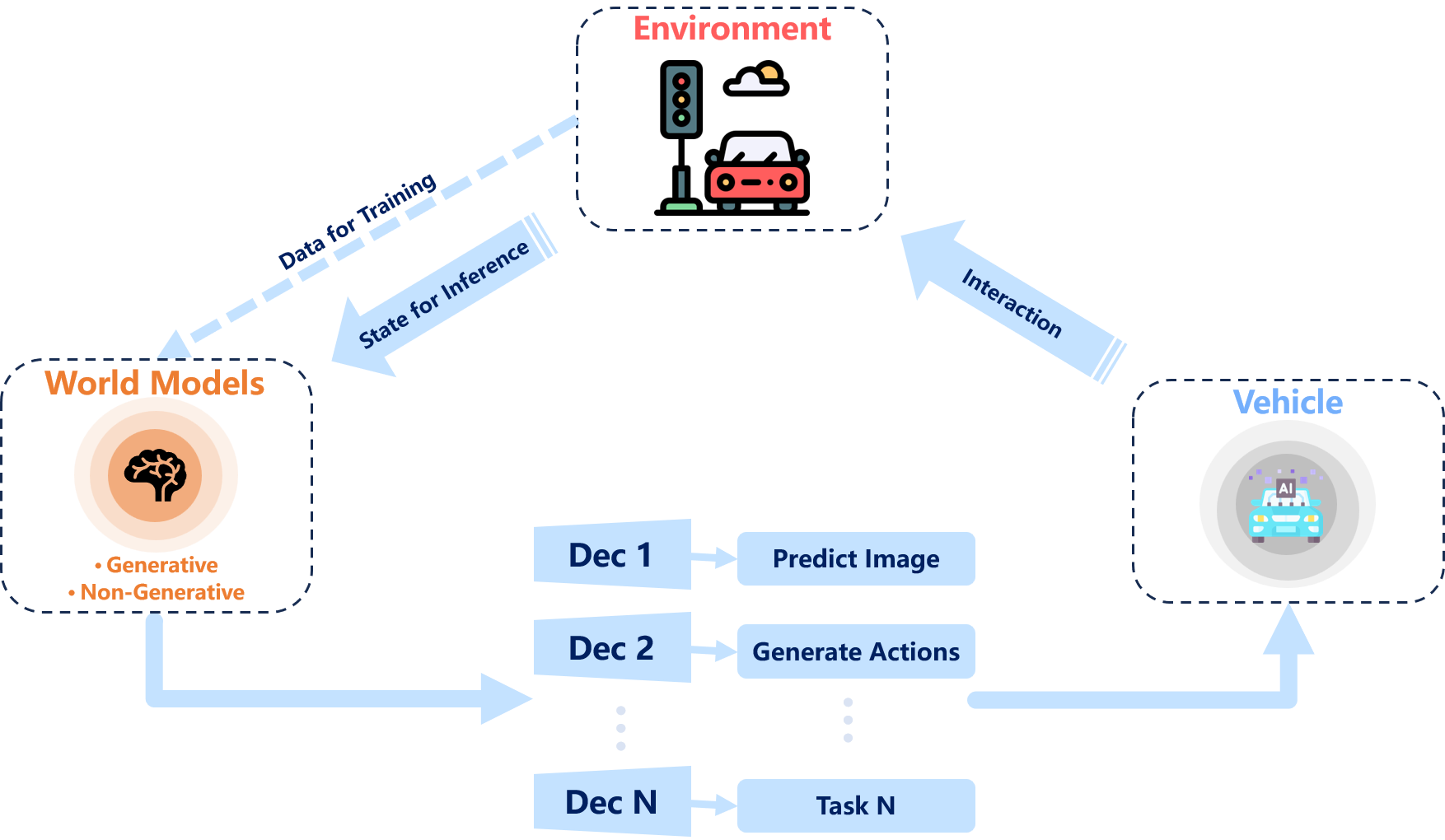}
    \caption{The pipeline diagram for enhancing autonomous driving with World Models. The World Models first learn the intrinsic evolutionary patterns by observing the traffic environment and then enhance autonomous driving by hooking up different decoders adapted to different driving tasks.}
    \label{fig:5}
\end{figure}

It can be seen that World Model is highly related to reinforcement learning, imitation learning, and deep generative models. However, utilizing World Models in reinforcement learning and imitation learning generally requires labeled data, and both SEM2 and MILE approaches mentioned are conducted within a supervised paradigm. There have also been attempts to combine reinforcement learning and unsupervised learning (UL) based on the limitations of labeled data\cite{144,145}. Due to the close relationship with self-supervised learning, deep generative models have become more and more popular, and researchers in this field have made many attempts. In the following, we will mainly review the exploratory applications of generative World Models in autonomous driving, the pipeline is illustrated in Fig.~\ref{fig:5}, 4.1 introduces the principles of various types of deep generative models and their applications in generative driving scenarios, 4.2 introduces the applications of generative World Models in autonomous driving, and 4.3 will introduce a class of non-generative methods.  

\subsection{Deep Generative Models}

Deep generative models generally include ariational autocoders (VAEs) \cite{146,147}, generative adversarial networks (GANs)\cite{26,148}, flow models\cite{149,150}, and autoregressive models (ARs)\cite{8,151,152}.

VAEs combine the ideas of self-encoders and probabilistic graphical models to learn underlying data structures and generate new samples. Rempe et al.\cite{153} used VAE to learn prior distributions of traffic scenarios and simulate the generation of accident-prone scenarios. GANs consist of generator and discriminator, which compete and enhance each other utilizing adversarial training, to ultimately achieve the goal of generating realistic samples. Kim et al.\cite{154} used a GAN model to observe sequences of unlabeled video frames and their associated action pairs to simulate a dynamic traffic environment. The flow models generate similar data samples by transforming simple prior distributions into complex posterior distributions through a series of invertible transformations. Kumar et al.\cite{155} used the flow model to achieve multi-frame video prediction. ARs are a class of sequence analysis methods, based on the autocorrelation between the sequence data, describing the relationship between the present and the past, and the estimation of the model parameters is usually done leveraging the least squares method and maximum likelihood estimation. For example, GPT uses maximum likelihood estimation for model parameter training. Feng et al.\cite{156} achieved the generation of future trajectories of vehicles based on autoregressive iterations. The diffusion model is a typical autoregressive method that learns the process of gradual denoising from purely noisy data. With its strong generative performance, the diffusion model is the new SOTA among current deep generative models. Works such as\cite{157,158,159} demonstrated that the diffusion model has a strong ability to understand complex Scenarios, and the video diffusion model can generate higher quality videos. Work such as\cite{160,161} utilized the diffusion model to generate complex and diverse driving scenarios.

\subsection{Generative Methods}

Based on the powerful capabilities of deep generative models, leveraging deep generative models as World Models to learn driving scenarios to enhance autonomous driving has become popular. The following section will review the applications of leveraging deep generative models as World Models in autonomous driving. In Table \ref{tab:2} we provide a brief overview of some representative works.

\textbf{Point Cloud involved Models.} Zhang et al.\cite{165} built on Maskgit\cite{166} and recast it into a discrete diffusion model for point cloud prediction. This method utilized VQ-VAE\cite{167} to tokenize the observation data for label-free learning. Karlsson et al.\cite{162} used a hierarchical VAE to construct World Model, used latent variable prediction and adversarial modeling to generate pseudo-complete states, matched partial observations with pseudo-complete observations to predict future states, and evaluated it on the KITTI-360\cite{163} dataset. In particular, it utilized pre-trained vision-based semantic segmentation models to infer from raw images. Bogdoll et al\cite{bogdoll2023muvo} constructed a multi-modal autonomous generative World Model, MUVO, leveraging raw images and LiDAR data to learn a geometric representation of the world. And conditioned on actions, this mode achieved 3D occupancy prediction and can be directly applied to downstream tasks (e.g., planning). Similarly, Zheng et al.\cite{zheng2023occworld} used VQ-VAE to tokenize the 3D occupancy scene and constructed a 3D occupancy apace to learn a World Model that can predict the motion of the ego-vehicle and the evolution of the driving scenario. To obtain finer-grained scene information, Min et al.\cite{min2023uniworld} used unlabeled image-LiDAR for pre-training to construct World Model that can generate 4D geometric occupancy. 

\textbf{Image-based Models.} To address the challenges of predicting future changes in driving scenarios, Wayve proposed a generative World Model, GAIA-1\cite{7}. GAIA-1 used transformer as World Model learned and predicted the next states of the input video, text, and action signals, and then generated realistic driving scenarios. For the learning of video streams, GAIA-1 adopted self-supervised learning, which can learn scaled data and obtain a comprehensive understanding of driving scenarios. Wang et al.\cite{164} devised a two-stage training strategy. Initially, a diffusion model was employed to learn driving scenarios and gain an understanding of structured traffic information. Subsequently, a video prediction task was used to construct a world model, designated DriveDreamer. Notably, by integrating historical driving behaviors, this approach enables the generation of future driving actions. Zhao et al.\cite{zhao2024drivedreamer} constructed DriveDreamer-2 on top of the DriveDreamer framework by integrating LLM, which generates the corresponding agent trajectories based on user descriptions, and HDMap information to controllably generate driving videos. Wang et al.\cite{wang2023driving} generated the driving videos by jointly modeling the future multi-views and multi-frames. This approach greatly improved the consistency of the generated results, and end-to-end motion planning was generated based on this.

In the industry, at the 2023 CVPR Autonomous Driving Workshop, Tesla researcher Ashok Elluswamy presented their work in utilizing generative large model\cite{168} to generate future driving scenarios. In the demonstration, it was seen that the videos generated by Tesla's generative large model were very close to those captured from real vehicles. It also can generate annotation-like semantic information, indicating that the model also has some semantic-level understanding and reasoning capabilities. Tesla named their work ``Learning a General World Model" and it can be seen that their understanding is to build a generalized World Model. By learning from a large amount of visual data captured from real vehicles, Tesla intends to build a large-scale FM for autonomous driving, which can understand the dynamic evolution of the world.

\textbf{Visual Prediction.} Vision is one of the most direct and effective means by which humans acquire information about the world because the feature information contained in image data is extremely rich. Numerous previous works\cite{139,140,141,145,169} have accomplished the task of image generation through World Model, demonstrating that World Model has a good understanding and reasoning ability for image data. However, these are mainly focused on image generation and are still lacking in video prediction tasks that can better represent the dynamic evolution of the world. Video prediction tasks require a deeper understanding of world evolution and also stronger guidance for downstream tasks. In the research works\cite{7,162}, they all effectively predicted generated future traffic scenarios, where self-supervised learning may be key. Previous work has explored this as well. Villegas et al.\cite{170} trained a model leveraging raw images and proposed a hierarchical long-term video prediction method combining low-level pixel space and high-level feature space (e.g., landmarks), achieving longer video prediction compared to the work\cite{141}. Endo et al.\cite{171} constructed a model under the self-supervised learning paradigm for predicting future traffic Scenarios from single-frame images to predict the future. Based on a denoising diffusion model with probabilistic conditional scores, Voleti et al.\cite{172} trained the model by randomly masking the past frames or future frames unlabeled, which allowed block-by-block autoregressions to generate videos of arbitrary length. Finn et al.\cite{173} proposed to physically interact with the world under unsupervised conditions and realize video prediction by predicting the distribution of pixel motion. Micheli et al.\cite{174} verified the effectiveness of leveraging an autoregressive transformer as a World Model, and achieved the prediction of game images by training the parameters through self-supervised learning. Wu et al.\cite{175} constructed an object-centered World Model to learn complex spatio-temporal interactions between objects and generated high visual quality future prediction. 

Inspired by LLM, Wang et al.\cite{wang2024worlddreamer} consider world modeling as unsupervised visual sequence modeling. The visual input is mapped into discrete tokens using VQ-GAN\cite{esser2021taming}, and then the Spatio-Temporal Transformer is used to predict the masked tokens to learn the physical evolutionary patterns in them, thus gaining the ability to generate videos in various scenarios. Analogous to LLM's tokens, OpenAI researchers transformed visual data into patches to propose the video generation model Sora. To address the high-dimensionality of visual data, they compressed the visual data into a lower-dimensional latent space and then generated a latent representation in this latent space through diffusion. This representation was then mapped back to the pixel space to realize video generation. By learning from Internet-scale data, Sora realizes the scaling law in the video domain, and Sora can generate coherent high-definition videos based on diverse prompts. In the same year, Google proposed Genie\cite{bruce2024genie}, a generative interactive model that uses unlabeled Internet gaming videos for training. In particular, Genie proposed a latent action model to infer latent actions between each frame and constructed a codebook for latent actions through training. To utilize the model, the user selects the initial frame and the specified latent action and autoregressively generates future frames. As the model size and batch size increase, Genie also demonstrates scaling results. In contrast, Sora is designed to generate video content with high fidelity, variable duration, and resolution. While not as advanced in video quality as Sora, Genie is optimized for building generative interactive environments in which the user can manipulate frame-by-frame to generate video.

The preceding studies demonstrate the efficacy of World Models in enhancing autonomous driving. World Models can be directly embedded into autonomous driving models to accomplish various driving tasks. Furthermore, there are explorations of learning to build general World Models from large-scale visual data, such as Sora and Genie. These FMs can be utilized for data generation (to be discussed in Section 5). In addition, based on FMs' generalization ability, they can be employed to perform a multitude of downstream tasks, or even be utilized to simulate the world.

\begin{center}
\begin{longtable}[h]{
         >{\raggedright\arraybackslash}m{0.4cm}
         >{\raggedright\arraybackslash}m{1.3cm}
         >{\raggedright\arraybackslash}m{1.3cm}
         >{\raggedright\arraybackslash}m{1.2cm}
         >{\raggedright\arraybackslash}m{1.2cm}
         >{\raggedright\arraybackslash}m{7.6cm}
        }
    \caption{Works on the use of World Models for prediction}
    \label{tab:2} \\
    
            \hline
            &
            Authors & Input & Output & Learning & Description \\
            \hline
            
            \multirow{16}{.4cm}{AD} & 
            \makecell[l]{Karlsson\\et al.\cite{162}\\2023} & Images, Point Clouds & Semantic Point Clouds & SSL & This work utilizes a hierarchical VAE to construct World Model and generates pseudo-complete states and matches them with partial observations to predict future states.\\
            \cline{2-6}
            & \makecell[l]{Hu\\et al.\cite{7}\\2023} & Videos, Text, Actions & Videos & SSL & This work utilizes an autoregressive transformer to construct World Model and leverages DINO, a self-supervised image model, to tokenize images.\\
            \cline{2-6}
            & \makecell[l]{Wang\\et al.\cite{164}\\2023} & Images, HDMap, 3D Box, Text, Actions & Videos, Actions & SSL & This work obtains comprehension of the structured traffic information. Then the prediction is formalized into a generative probabilistic model.\\
            \cline{2-6}
            & \makecell[l]{Zhang\\et al.\cite{165}\\2023} & Point Clouds, Actions & Point Clouds & UL/SSL & This work utilizes a discrete diffusion model for point cloud prediction, which is a spatio-temporal transformer. And this work leverages VQ-VAE to tokenize sensor observations.\\
            \cline{2-6}
            & \makecell[l]{Zheng\\et al.\cite{zheng2023occworld}\\2023} & 3D Occupancy Scene & Scene, Ego-vehicle Motion & SSL & By constructing a 3D occupancy space, a World Model is trained to predict the next scene from previous Scenarios, following an autoregressive manner. This work utilizes VQ-VAE for discretizing the 3D occupancy scene into tokens.\\
            \hline

            \multirow{5}{.4cm}{AD} &
            \makecell[l]{Min\\et al.\cite{min2023uniworld}\\2023} & Image-LiDAR pairs & 4D GO & UL/SSL & This work proposes a spatial-temporal World Model for unified autonomous driving pretraining.\\
            \cline{2-6}
            & \makecell[l]{Bogdoll\\et al.\cite{bogdoll2023muvo}\\2023} & Actions, Point Clouds, Images & Point Clouds, Images, 3D OG & UL/SSL & This work leverages raw data to learn a sensor-agnostic 3D occupancy representation and predicts future states conditional on actions.\\
            \hline
            
            \multirow{8}{.4cm}{VP} &  
            \makecell[l]{Finn\\et al.\cite{173}\\2016} & Videos & Videos & UL/SSL & This work proposed to interact with the world under unsupervised conditions and develops an action-conditioned model for video prediction.\\  
            \cline{2-6}
            & \makecell[l]{Wu\\et al.\cite{175}\\2022} & Videos & Videos & UL/SSL & This work leverages a pre-trained object-centric model to extract object slots from each frame. These slots are then forwarded to a transformer and used to predict future slots. \\
            \cline{2-6}
            & \makecell[l]{Wang\\et al.\cite{wang2024worlddreamer}\\2024} & Images, Videos, Text, Actions & Videos & SSL & Visual inputs are mapped into discrete tokens using VQ-GAN, and then the masked tokens are predicted using Transformer.\\ 
            \hline
\multicolumn{6}{l}{\makecell[l]{\small $\bullet$ This ``AD'' refers to ``Autonomous Driving'' and this ``VP'' refers to ``Visual Prediction''.\\ \small $\bullet$ This ``GO'' refers to ``Geometric Occupancy'' and this ``OG'' refers to ``Occupancy Grids''.}} 

\end{longtable}
\end{center}

\subsection{Non-Generative Methods}

\begin{figure}[h]
    \centering
    \includegraphics[scale=0.42]{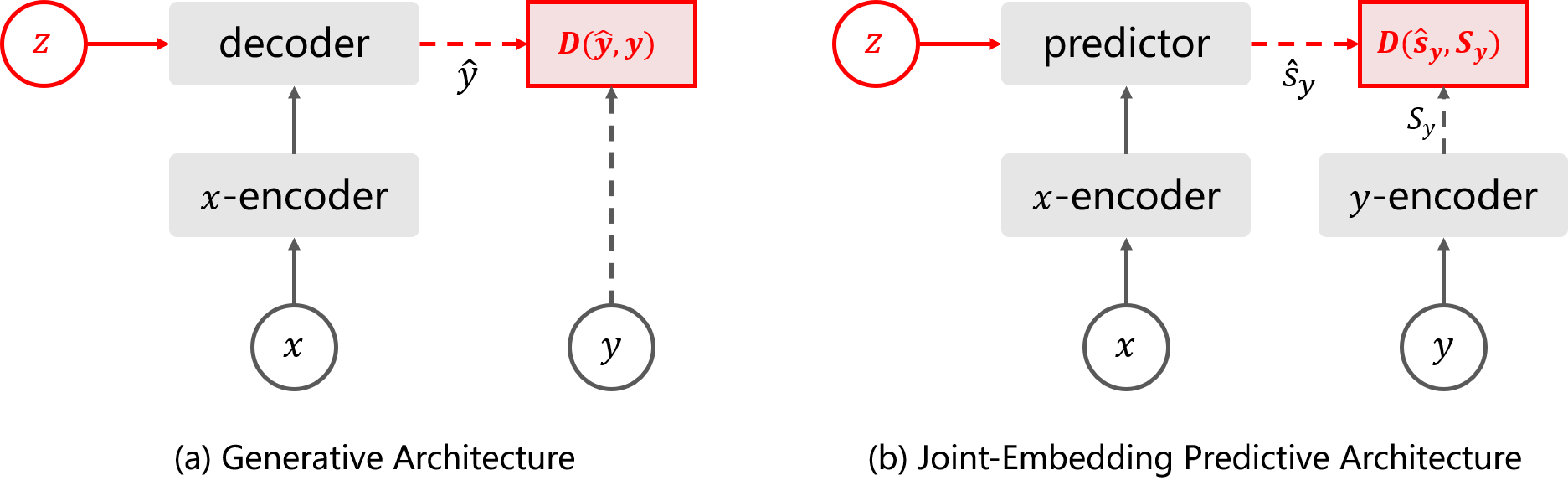}
    \caption{Comparison of the architecture of generative and non-generative methods\cite {176}.}
    \label{fig_image3}
\end{figure}

In contrast to generative World Models, LeCun\cite{135} elaborated on different conceptions of World Model by proposing a Joint Extraction and Prediction Architecture (JEPA) based on energy-based model. This is a non-generative and self-supervised architecture, as it does not predict the output $y$ directly from the input $x$, but encodes $x$ as $sx$ to predict $sy$ in representation space, as illustrated in Fig.\ref{fig_image3}. This has the advantage that it does not have to predict all the information about $y$ and can eliminate irrelevant details.

Since its proposal, the JEPA architecture has been applied by several scholars in different domains with excellent performance. In the graph domain, Skenderi et al.\cite{178} proposed Graph-JEPA, which is a JEPA model for graph domains. It divides the input graph into subgraphs and then predicts the representation of the target subgraph in the context subgraph. Graph-JEPA has obtained excellent performance in both graph classification and regression problems. In the field of audio, Huang et al.\cite{fei2023jepa} proposed A-JEPA, which applies the mask modeling principle to audio. Following experimental validation, A-JEPA has been demonstrated to perform well in speech and audio classification tasks. Sun et al. proposed JEP-KD\cite{sun2024jep}, which employs an advanced knowledge distillation method to enhance the effectiveness of Visual Speech Recognition (VSR) and narrow the performance gap between VSR and Automatic Speech Recognition (ASR).

In the field of computer vision, Bardes et al.\cite{177} proposed MC-JEPA, which employs the JEPA architecture and a self-supervised learning approach to facilitate co-learning of optical flow and content features, thereby enabling the acquisition of dynamic content features. From video, MC-JEPA performs well in a variety of tasks, including estimation of optical flow, and segmentation of images and videos. META\cite{176} proposed I-JEPA for learning highly semantic image representations without relying on manual data enhancement. The combination of I-JEPA with Vision Transformers yielded strong downstream performance in a variety of tasks, including linear classification, object counting, and depth prediction. Building on I-JEPA, META applies JEPA to the video domain by proposing V-JEPA\cite{bardes2024revisiting}. This method combines mask prediction with the JEPA architecture to train a series of V-JEPA models with feature prediction as the goal of self-supervised learning. Experimental results demonstrate that these models exhibit excellent performance in a range of computer vision downstream tasks, including action recognition, action classification, and target classification.

To date, no literature has been identified that directly applies the JEPA to the field of autonomous driving. Nevertheless, it has great potential. Firstly, instead of predicting the video in pixel space, non-generative world models make feature predictions in representation space. This eliminates many irrelevant details. For example, in the scene prediction task of autonomous driving, we are more interested in the future movements of other traffic participants on the current road. Furthermore, for other vehicles that are not on the current road of the autonomous vehicle, for example, situated next to an elevated road parallel to the current road, we do not consider their future motion trajectories. The JEPA model eliminates these irrelevant details and reduces the complexity of the problem. Additionally, V-JEPA has demonstrated its ability to learn features in video. By analyzing a sufficiently large number of driving videos, it is anticipated that V-JEPA will be widely used in tasks such as generating driving scenarios and predicting future environmental states.

\section{Data Augmentation Based on Foundation Models}

As deep learning continues to evolve, the performance of FMs with pre-training and fine-tuning as the underlying architecture is improving. FMs are spearheading the transition from rule-driven to data-driven learning paradigms. The significance of data as a key aspect of model learning is evident. A substantial quantity of data is utilized in the training process of an autonomous driving model to facilitate the model's comprehension and decision-making abilities in diverse driving scenarios. Nevertheless, the collection of realistic data is a time-consuming and laborious process, so data augmentation is crucial to improving the generalization ability of automatic driving models. 

The realization of data augmentation needs to consider two aspects: on the one hand, how to obtain large-scale data so that the data fed to the autonomous driving system is diverse and extensive; on the other hand, how to obtain as much high-quality data as possible so that the data which is used to train and test the autonomous driving models is accurate and reliable. Related works have also roughly chosen two directions to enhance the autonomous driving data, one is to enrich the data content of existing datasets and enhance the data features of driving scenarios, and the other is to generate driving scenarios with multiple levels through simulation. In the following, a review of related works on enhancing data based on FMs will be presented, in Section 5.1 we describe related work on extending datasets, and in Section 5.2 we describe related work on generating driving scenarios. Table \ref{tab:5} provides a brief overview of some representative works.

\subsection{Expansion of Autonomous Driving Datasets}

Existing autonomous driving datasets are mostly obtained by recording sensor data and then labeling the data. The features of the data obtained in this are usually low-level and exist more at the level of numerical representation, which is insufficient for the visuo-spatial features of the autonomous driving scenarios. Natural language descriptions are seen as an effective way to enhance scene representation\cite{4}, Flickr30k\cite{108}, RefCOCO\cite{109}, RefCOCOg\cite{110}, and CLEVR-Ref\cite{111} use concise natural language descriptions to identify the corresponding visual regions in an image. Talk2Car\cite{112} fused image, radar, and LiDAR data to construct the first object-referenced dataset containing language commands for self-driving cars. However, the Talk2Car dataset allowed only one object to be referenced at a time. CityFlow-NL\cite{113} constructed a dataset for multi-target tracking through natural language descriptions, and Refer-KITTI\cite{114} achieved prediction of arbitrary target tracking by leveraging natural language queries in the corresponding task.

FMs provide new ideas for enriching and expanding autonomous driving datasets under their advanced semantic understanding, reasoning, and interpretation capabilities. Qian et al.\cite{115} created NuScenes-QA, a visual question-and-answer dataset for autonomous driving in 3D multi-view driving scenarios, by encoding question descriptions through a language model and obtaining answers through feature fusion with sensor data. Significant progress was made in the use of natural language prompts. Wu et al.\cite{116} extended NuScenes-QA by constructing the dataset NuPrompt by capturing and combining natural language elements and then invoking LLM to generate the descriptions. The dataset provided a finer match between the 3D instances and each of the prompts, which helped to characterize objects in the autopilot images more accurately. Sima et al.\cite{drivelm} took into account the interactions of the traffic elements and constructed Graph Visual Question Answering by extending the nuScenes dataset\cite{128} with BLIP-2, which can better clarify the logical dependencies between objects and the hierarchy of driving tasks. In addition to directly extending the augmented autonomous dataset, some scholars have also integrated the Chain-of-Thought (CoT) capability of LLM and the cross-modal capability of the vision model to build an automatic annotation system, OpenAnnotate3D\cite{117}, which can be used for multi-modal 3D data. Expanding the dataset by utilizing the advanced understanding, reasoning, and interpretation capabilities of the underlying models can help to better assess the interpretability and control of the autonomous driving system, thus improving the safety and reliability of the autonomous driving system. A comparison of some representative work is shown in Table \ref{tab:3}.

\begin{table}[h]
\caption{Comparison of extended datasets. `-' means unavailable.}
\label{tab:3}
\resizebox{1.0\linewidth}{!}{
\begin{tabular}{l|c|c|c|c|c|c|c|c}
\toprule[1pt]
Dataset           & Source    & Based FMs & Modality & 3D & Multi-Views & VIdeos & Frames & QA Pairs \\ \toprule[1pt]
RefCOCO\cite{110}       & COCO      & None & \makecell{image \\ referring expression}  & ×  & × & - & 26711  & -  \\ \hline
Refer-KITTI\cite{114}   & KITTI     & None & \makecell{image \\ point cloud \\ object referral} & ×  & × & 18 & 6650   & -  \\ \hline
Talk2Car\cite{112}      & nuScences & None & \makecell{image \\ point cloud \\ driving command} & √  & ×  & -  & 9217   & -  \\ \toprule[1pt]
nuPrompt\cite{116}      & nuScences & GPT-3.5 & \makecell{image \\ point cloud \\ Question-Answering} & √  & √ & 850 & 34149  & 35k \\ \hline
DriveLM-nuScences\cite{drivelm} & nuScences & BLIP-2 & \makecell{image \\ point cloud \\ Question-Answering} & √  & √ & - & 4871   & 443k \\ \toprule[1pt]
\end{tabular}
}
\end{table}

\subsection{Generation of Driving Scenarios}

The diversity of driving scenarios is of great significance for autonomous driving. To obtain a better generalization ability, autonomous driving models must learn a wide variety of scenarios. However, the reality is that driving scenarios conform to a long-tailed distribution (It is a probability distribution in which a significant proportion of the observations or instances are concentrated in the tail(s) of the distribution, away from the center or mean.). \emph{The ``long-tail problem'' of autonomous driving vehicles is that they are capable of handling situations that are frequently encountered, but are unable to cope with corner cases in rare or extreme situations.} To address the long-tail problem, the key is to get as many corner cases as possible. Nevertheless, it is inefficient to limit the collection to real scenarios. For instance, in CODA\cite{li2022coda}, work on corner case mining, there are only 1,057 valid data out of 1 million data.

Given the above, the generation of large-scale and high-quality driving scenario data necessitates the capacity to actively create a multitude of driving scenarios. Traditional methodologies may be classified into two primary categories: rule-based and data-driven. Rule-based approaches, as exemplified by the literature cited in references\cite{118,119,120,121}, necessitate the utilization of predefined rules, are inadequate for the characterization of complex environments, simulate simpler environments, and exhibit limited generalization ability. In contrast, data-driven approaches\cite{122,123,124,125} utilize driving data to train the model, enabling it to continuously learn and adapt. However, data-driven approaches often necessitate a substantial quantity of labeled data for training, impeding further development of driving scenario generation. Additionally, this approach lacks control and is unsuitable for custom generation. Recently, FMs have achieved considerable success, and the generation of higher-quality driving scenarios through FMs has also attracted significant research attention. On the one hand, the diversity and accuracy of data generation can be enhanced based on the powerful understanding and reasoning capabilities of FMs. On the other hand, diverse prompts can be designed for controlled generation.

\begin{table}[h]
\centering
\caption{Video generation performance on nuScenes dataset. `-' means unavailable. The FID indicator and FVD indicator provide feedback on the image and video quality, respectively.}
\label{tab:4}
\begin{tabular}{l|c|c|c|c|c}
\toprule[1pt]
Method                & Based FMs  & Multi-View & Multi-Frame & FID↓ & FVD↓ \\ \toprule[1pt]
BEVGen\cite{swerdlow2024street} & None  & √   & ×  & 25.54 & -    \\ \cline{1-2} 
DriveGAN\cite{kim2021drivegan}  & None  & √   & √  & 73.4 & 502.3 \\ \toprule[1pt]

MagicDrive\cite{gao2024magicdrive} & CLIP  & √  & ×  & 16.20 & -  \\ \cline{1-2}
Panancea\cite{wen2023panacea} & CLIP  & √   & √  & 16.96 & 139   \\ \cline{1-2}

DriveDreamer\cite{164} & WM & √   & √  & 52.6 & 452.0 \\ \cline{1-2}
DriverDreamer-2\cite{zhao2024drivedreamer} & GPT-3.5 \& WM & √   & √  & 11.2 & 55.7 \\ \cline{1-2}
Driving-WM\cite{wang2023driving} & WM & √   & √  & 15.8 & 122.7 \\ \toprule[1pt] 
\end{tabular}
\end{table}

\textbf{Based on LLMs and VLMs.} In response to the fact that some long-tailed scenarios can never be collected in multi-view shots, Yang et al.\cite{130} fused verbal cues, BEV sketch, and multi-view noise to design a two-stage generative network BEVControl for synthesizing realistic street scene images. Nevertheless, BEVControl is insufficient for modeling foreground and background detail information. To address the difficulty of obtaining large-scale BEV representations, Li et al.\cite{131} proposed a spatio-temporal consistent diffusion framework, DrivingDiffusion, to autoregressively generate realistic multi-view videos controlled by 3D layouts. The quality of the generated data can be effectively enhanced by introducing local cue inputs into the vision model. For controllable generation, Wen et al.\cite{wen2023panacea} integrated text prompts, image conditions, and BEV sequences to design a controllable module to improve the controllability of driving scenarios generation. Gao et al.\cite{gao2024magicdrive} designed 3D geometric control by integrating text prompts with camera pose, road map, and object box fusion control to generate diverse road scenarios.

Based on the powerful understanding and reasoning ability of LLMs and VLMs, it has also become a research hotspot to embed them directly or guide the model to generate driving scenarios. Marathe et al.\cite{129} efficiently generated a dataset comprising 16 weather extremes via prompting leveraging a VLM. Nevertheless, the model had some extension limitations due to the phenomenon of pre-selected fixation in data selection. Chen et al.\cite{132} realized the combination of numerical vector modality and natural language by pairing control commands collected by reinforcement learning intelligence and question answers generated by LLM to directly construct new data. Zhong et al.\cite{133} proposed a scenario-level diffusion-based language-guided traffic simulation model, CTG++, which can generate instruction-compliant, realistic, and controllable traffic scenarios. Wang et al.\cite{105} utilized natural language descriptions as conceptual representations that were integrated with LLM to enrich the complexity of the generated scenarios by leveraging their powerful common-sense reasoning capabilities. The behavior of human drivers is also an important part of driving scenarios, Jin et al.\cite{134} proposed SurrealDriver, a generative driving agent simulation framework for urban environments based on LLM. By analyzing and learning from real driving data, SurrealDriver can capture the driver's behavior patterns and decision-making processes and generate behavior sequences that are similar to those in real driving. 

\textbf{Based on World Models.} To achieve the controllable generation of driving scenarios, Wang et al.\cite{164} combine text prompts and structured traffic constraints to guide the generation of pixel points with text descriptions. To obtain more accurate dynamic information, Wang et al.\cite{wang2023driving} incorporate driving actions into a controllable architecture, utilizing text descriptors, layouts, and ego actions to control video generation. However, these approaches introduce more structural information, which limits the interactivity of the model. To address this issue, Zhao et al.\cite{zhao2024drivedreamer} propose a novel approach that combines LLM with World Model. This approach involves using LLM to convert user queries into agents' trajectories, which are then used to generate HDMap. This HDMap then guides the generation of driving videos.

Efficient and accurate controllability generation can be achieved using FMs for driving scenarios. This will be able to provide diverse training data, which is important for improving the generalization ability of autonomous driving systems. A comparison of some representative work is shown in Table \ref{tab:4}. Furthermore, the generated driving scenarios can be used to evaluate different autonomous driving models to test and validate their performance. Of course, we should also be able to see that with the emergence of various large-scale FMs such as Sora and Genia, there are new potential ideas for the generation of autonomous driving videos. The models are not restricted to the driving domain but can be used for transfer learning utilizing models obtained from training in the general vision domain. While the current state of technology in this domain remains imperfect, we believe that in the future, with the breakthrough of related technologies, we can even use them to generate the various driving scenarios we need, and truly learn a World Model to simulate the world.

\begin{center}
\begin{longtable}[h]{
         >{\raggedright\arraybackslash}m{1.3cm}
         >{\raggedright\arraybackslash}m{3.7cm}
         >{\raggedright\arraybackslash}m{3.8cm}
         >{\raggedright\arraybackslash}m{3.0cm}
         >{\raggedright\arraybackslash}m{1.2cm}
        }
   
    \caption{Works on Data Augmentation}
    \label{tab:5} \\
            \hline
            & Authors & Input & Output & Learning\\
             \hline
             
            \multirow{4}{1.5cm}{Expand Dataset}
            & \makecell[l]{Qian et al.\cite{115} 2023} & \makecell[l]{Images, Text,\\Point Clouds} & \makecell[l]{Q-A Pairs} & \makecell[c]{SL}\\
             \cline{2-5}
            & \makecell[l]{Wu et al.\cite{116} 2023} & Images, Text & Object-Text Pairs & \makecell[c]{SL}\\
             \cline{2-5}
            & \makecell[l]{Zhou et al.\cite{117} 2023} & Images, Text & Labeled Data & \makecell[c]{SL}\\
             \hline
            
            \multirow{7}{1.5cm}{Generate Scenarios}
            & \makecell[l]{Marathe et al.\cite{129} 2023} & \makecell[l]{Objects, Scenarios,\\Weather Condition} & Multi-Weather Images & \makecell[c]{SL}\\
             \cline{2-5}
            & \makecell[l]{Yang et al.\cite{130} 2023} & Text, BEV Sketch, Multi-View Noise & Street-View Images & \makecell[c]{SL}\\
             \cline{2-5}
            & \makecell[l]{Li et al.\cite{131} 2023} & \makecell[l]{Layouts, Frames,\\Optical Flow Prior} & Multi-View Videos & \makecell[c]{SL}\\
             \cline{2-5}
            & \makecell[l]{Wen et al.\cite{wen2023panacea} 2023} & \makecell[l]{Text, BEV Sequence} & Multi-View Videos & \makecell[c]{SL}\\
             \cline{2-5}
            & \makecell[l]{Chen et al.\cite{132} 2023} & Objects, Text & Trajectory & \makecell[c]{SL}\\
             \hline
             
             \multirow{6}{1.5cm}{Generate Scenarios}
            & \makecell[l]{Zhong et al.\cite{133} 2023} & Text, Scenarios with Noise & Traffic Scenarios & \makecell[c]{SL}\\
             \cline{2-5}
            & \makecell[l]{Wang et al.\cite{105} 2023} & Images, Knowledge & Latent Space Simulation & \makecell[c]{SL}\\
             \cline{2-5}
            & \makecell[l]{Jin et al.\cite{134} 2023} & Text, Simulation of Urban Driving & Driving Maneuvers & \makecell[c]{SL}\\
             \cline{2-5}
            & \makecell[l]{Zhao et al.\cite{zhao2024drivedreamer} 2024} & Text & Videos & \makecell[c]{SL}\\
            \hline
\multicolumn{5}{l}{\small $\bullet$ This ``Q-A'' refers to ``Question-Answer''.}
\end{longtable}
\end{center}

\section{Conclusion and Future Directions}

This paper provides a comprehensive overview of the application of FMs to autonomous driving. In Section 3, recent works on the application of FMs such as LLMs and VLMs to autonomous driving are summarized in detail. In Section 4, we present an exploratory application of the World Models to the field of autonomous driving. In Section 5, recent works on data augmentation of the FMs are detailed. Overall, the FMs can effectively assist autonomous driving in terms of both augmenting the data and optimizing the model. 

To evaluate the effectiveness of FMs in autonomous driving, we compare different FMs with traditional methods in terms of their effectiveness in motion planning in Table \ref{tab:6}. Due to the relative maturity of LLMs and VLMs, it can be observed that methods based on them to enhance autonomous driving have been improved overall. In contrast, WMs-based approaches are still undergoing further exploration, with relatively less work published. Nevertheless, through the previous analysis, we can also see that World Models are excellent at learning the evolutionary laws of the physical world and have great potential for improving autonomous driving.

\begin{table}[h]
\caption{Motion planning performance on the nuScenes validation dataset. †: Results of perception and prediction from UniAD. ‡: Results of perception and prediction from VAD. *: Results of perception and prediction from dataset annotations.}
\centering
\label{tab:6}
\begin{tabular}{l|c|cccc|cccl}
\toprule[1pt]
\multirow{2}{*}{Method} & \multirow{2}{*}{Based FMs} & \multicolumn{4}{c}{L2(m)↓} & \multicolumn{4}{c}{Collision(\%)↓} \\ \cline{3-10}
              &      & 1s   & 2s   & 3s   & Avg. & 1s   & 2s   & 3s   & Avg. \\ \cline{1-10}
ST-P3\cite{hu2022st}         & None & 1.33 & 2.11 & 2.90 & 2.11 & 0.23 & 0.62 & 1.27 & 0.71 \\
UniAD\cite{104}         & None & 0.48 & 0.96 & 1.65 & 1.03 & 0.05 & 0.17 & 0.71 & 0.31 \\
VAD\cite{jiang2023vad}           & None & 0.41 & 0.70 & 1.05 & 0.72 & 0.07 & 0.17 & 0.41 & 0.22 \\
GenAD\cite{zheng2024genad}         & None & 0.36 & 0.83 & 1.55 & 0.91 & 0.06 & 0.23 & 1.00 & 0.43 \\ \toprule[1pt]

GPT-Driver\cite{72}†    & LLM  & 0.21 & 0.43 & 0.79 & 0.48 & 0.16 & 0.27 & 0.63 & 0.35 \\
GPT-Driver\cite{72}*    & LLM  & 0.20 & 0.42 & 0.72 & 0.44 & 0.14 & 0.25 & 0.60 & 0.33 \\

Agent-Driver\cite{mao2023language}†  & LLM  & 0.22 & 0.65 & 1.34 & 0.74 & 0.02 & 0.13 & 0.48 & 0.21 \\

DriveVLM-Dual\cite{tian2024drivevlm}† & VLM  & 0.17 & 0.37 & 0.63 & 0.39 & 0.08 & 0.18 & 0.35 & 0.20 \\
DriveVLM-Dual\cite{tian2024drivevlm}‡ & VLM  & 0.15 & 0.29 & 0.48 & 0.31 & 0.05 & 0.08 & 0.17 & 0.10 \\

VLP-UniAD\cite{pan2024vlp}†     & LLM  & 0.36 & 0.68 & 1.19 & 0.74 & 0.03 & 0.12 & 0.32 & 0.16 \\
VLP-VAD\cite{pan2024vlp}‡      & LLM  & 0.30 & 0.53 & 0.84 & 0.55 & 0.01 & 0.07 & 0.38 & 0.15 \\ \toprule[1pt]

OccWorld-O\cite{zheng2023occworld}*    & WM   & 0.43 & 1.08 & 1.99 & 1.17 & 0.07 & 0.38 & 1.35 & 0.60 \\
Drive-WM\cite{wang2023driving}      & WM   & 0.43 & 0.77 & 1.20 & 0.80 & 0.10 & 0.21 & 0.48 & 0.26 \\ \toprule[1pt]
\end{tabular}
\end{table}

\textbf{Challenges and Future Directions.} Nevertheless, it is evident from previous studies that FM-based autonomous driving is not yet sufficiently mature. This phenomenon can be attributed to several factors. FMs suffer from the problem of hallucination\cite{zhang2023siren,liu2024exposing}, as well as the fact that there are still limitations in learning video, a high-dimensional continuous modality. Additionally, deployment issues caused by inference latency\cite{pope2023efficiently,weng2023inference} and potential ethical implications and societal impact also be considered.

\textbf{Hallucination.} The hallucination error problem is mainly manifested as misrecognition in autonomous driving, such as wrong target detection, which may cause serious safety accidents. The phantom problem mainly arises because of the limited samples in the dataset or because the model is affected by unbalanced or noisy data, and the stability and generalization ability needs to be strengthened by utilizing expanding data and adversarial training. 

\textbf{Real-world deployment.} As previously discussed, the majority of current research on FMs in autonomous driving is based on open-source dataset experiments\cite{tian2024drivevlm,pan2024vlp} or closed-loop experiments in simulation environments\cite{lmdrive,wang2023drivemlm}, which is insufficient for real-time considerations. Additionally, some studies\cite{pope2023efficiently,weng2023inference} have highlighted that large models have certain inference latency, which could potentially lead to significant safety concerns in autonomous driving applications. To further explore the effectiveness of FMs for real-time applications in autonomous driving, we conducted an experimental\cite{ww}. We used LoRA to fine-tune the LLaMA-7B\cite{44}, and the fine-tuned LLM can reason to generate driving language commands. To verify its real-time performance in driving scenarios, we reasoned on a single GPU A800 and a single GPU 3080, respectively, and the time required to generate 6 tokens is 0.9s and 1.2s, effectively verifying that vehicle deployment of FM is possible. To further utilize the computing resources in the vehicle, other modules, such as the intelligent cockpit, can be considered to be deployed in the cloud. Therefore, the final form of deployment in the real world can be speculated: deploying the intelligent cockpit module in the cloud, while the intelligent driving module is directly deployed in the vehicle, is a relatively more reasonable choice. In the future, with the improvement of edge computing and in-vehicle computing capabilities\cite{xu2023sparks}, it may gradually move to a hybrid deployment model of vehicle, road, and cloud to further improve real-time response capabilities and privacy protection.

\textbf{AI Alignment.} The deepening of FMs into various industries, including autonomous driving, is a significant trend. Nevertheless, as related research continues, so do the risks to human society. Advanced AI systems exhibiting undesirable behaviors (e.g., spoofing) are a cause for concern, especially in areas such as autonomous driving, which is directly related to personal safety, and requires serious discussion and reflection. In response to this, AI Alignment has been proposed and is currently being developed. The objective of AI Alignment is to align the behaviors of AI systems with human intentions and values. This approach focuses on the goals of AI systems rather than their capabilities\cite{leike2018scalable}. AI Alignment facilitates the risk control, operational robustness, human ethicality, and interpretability of advanced AI systems when implemented in various domains\cite{ji2024ai}. This is a substantial body of research encompassing numerous AI-related disciplines. As this paper concentrates on the domain of autonomous driving and does not delve into the specifics of risk causes and solutions, we will not elaborate further here. In the field of autonomous driving, it is important to note that while promoting the application of FMs, researchers must establish reasonable technical ethics based on the guidance of AI Alignment. This includes paying attention to the issues of algorithmic fairness, data privacy, system security, and the human-machine relationship. Furthermore, it is essential to promote the unity of technological development and social values to avoid potential ethical and social risks.

\textbf{Visual Emergent Abilities.} FMs have seen amazing emergent abilities with model scaling and demonstrated success in natural language processing. Nevertheless, in the context of autonomous driving, this line of research faces additional open challenges due to limited available data and extended context length issues. These challenges contribute to an inadequate understanding of macroscopic driving scenarios, thereby complicating long-term planning in this field. Driving video is a high-dimensional continuous modal with an extremely large amount of data (several orders of magnitude larger compared to textual data). Hence, training large vision models requires a more macroscopic scene distribution to embed enough video frames to reason about complex dynamic scenarios, which requires a more robust network structure and training strategies to learn this information. Bai et al.\cite{bai2023sequential} proposed a two-stage approach in a recent study, in which images are converted into discrete tokens to obtain ``visual sentences", and then autoregressive predictions are made, similar to the standard approach for language model\cite{11}. Another promising solution may lie in World Models. As described in Section 4, World Models can learn the intrinsic evolutionary laws of the world by observing a small number of events that are either relevant or irrelevant to the task. However, World Models also have certain limitations in exploratory applications, where uncertainty in the predictive outcomes of the models, as well as learning what kind of data captures the intrinsic laws of how the world works, still warrant further exploration. 

In conclusion, although there are many challenges to be solved in applying Foundation Models to autonomous driving, its potential has already begun to take shape. In the future, we will continue to monitor the progress of Foundation Models applied to autonomous driving.

\section*{Acknowledgments}

\subsection*{Author Contributions} 
Hong Chen, Hongqing Chu, B.G., and J.W. conceived and designed the study. J.W., J.G., and J.Y. wrote the manuscript draft. X.C., Y.C., Q.Y., and H.E.T gave some helpful suggestions. Hong Chen, Hongqing Chu, B.G., and J.C. revised the manuscript. The authors read and approved the final manuscript.

\subsection*{Funding}
This work was supported in part by the National Nature Science Foundation of China (No. 62373289, 62273256, 62088101), and the Fundamental Research Funds for the Central Universities. 

\subsection*{Conflicts of Interest}
The authors declare that there is no conflict of interest regarding the publication of this article.

\printbibliography
\end{CJK}
\end{document}